\pdfoutput=1

\documentclass[11pt]{article}

\usepackage{acl}

\usepackage{times}
\usepackage{latexsym}
\usepackage{amsfonts} 
\usepackage{enumitem}
\usepackage{booktabs}
\usepackage{multirow}
\usepackage{array}
\usepackage{graphicx}
\usepackage{amsmath}
\usepackage{amssymb}
\usepackage{amsfonts}
\usepackage{subcaption}
\usepackage{makecell}
\usepackage{color,soul}
\usepackage{algorithm}
\usepackage[noend]{algpseudocode}
\usepackage{xcolor,colortbl}
\usepackage{xspace}
\usepackage{CJKutf8} 
\usepackage[export]{adjustbox} 

\DeclareMathOperator*{\argmax}{argmax} 
\DeclareMathOperator*{\argtopk}{argTopk} 
\newcolumntype{L}[1]{>{\PreserveBackslash\raggedright}p{#1}}
\newcolumntype{R}[1]{>{\raggedleft\let\newline\\\arraybackslash\hspace{0pt}}m{#1}}

\definecolor{Gray}{gray}{0.85}
\newcolumntype{a}{>{\columncolor{Gray}}r}

\usepackage{url}
\usepackage{hyperref}

\definecolor{green}{rgb}{0.1,0.1,0.1}
\definecolor{gitgreen}{HTML}{006400}
\definecolor{chocolate}{HTML}{D2691E}
\definecolor{maroon}{HTML}{A00000}
\definecolor{indigo}{HTML}{4B0082}
\definecolor{green}{HTML}{008000}

\definecolor{myred}{rgb}{0.8, 0.0, 0.0}
\definecolor{myblue}{rgb}{0.2, 0.2, 0.6}

\usepackage[T1]{fontenc}

\usepackage[utf8]{inputenc}

\usepackage{microtype}

\makeatletter
\newcommand*\myfontsize{%
  \@setfontsize\myfontsize{7.7}{9}%
}
\makeatother

\newcommand{\Np}{Nonparametric}
\newcommand{\np}{nonparametric}
\newcommand{\ours}{\textsc{NpM}}
\newcommand{\ourssingle}{\textsc{NpM single}}
\newcommand{\knn}{$k$NN}
\newcommand{\knnlm}{$k$NN-LM}
\newcommand{\mask}{\texttt{[MASK]}}
\newcommand{\maskS}{\texttt{[MASK$_\mathrm{s}$]}}
\newcommand{\maskE}{\texttt{[MASK$_\mathrm{e}$]}}
\newcommand{\ngram}{$n$-gram}
\newcommand{\ngrams}{$n$-grams}

\newcommand{\phrase}{phrase} 
\newcommand{\phrases}{phrases}

\newcommand{\templama}{TempLAMA$_{19}^{22}$ }
\newcommand{\templamabold}{\textbf{TempLAMA}$_\mathbf{19}^\mathbf{22}$ }

\usepackage{mathtools}

\newcommand{\mybrown}[1]{\textcolor{brown}{#1}}
\newcommand{\myorange}[1]{\textcolor{orange}{#1}}
\newcommand{\mygreen}[1]{\textcolor{green}{#1}}

\definecolor{cadmiumgreen}{rgb}{0.0, 0.42, 0.24}
\newcommand{\mydarkgreen}[1]{\textcolor{cadmiumgreen}{#1}}
\newcommand{\myindigo}[1]{\textcolor{indigo}{#1}}
\newcommand{\best}[1]{\textbf{\mygreen{#1}}}
\PassOptionsToPackage{svgnames,x11names,dvipsnames}{xcolor}
\usepackage[most]{tcolorbox}

\newtcbox{\inlinebox}[1][]{enhanced,
 box align=base,
 nobeforeafter,
 colframe=Green4,
 size=small,
 left=-5pt,
 right=-5pt,
 boxsep=0pt,
 #1}

\usepackage{tikz}
\usetikzlibrary{tikzmark}
\newcommand{\mytextbox}[2]{\tikzmarknode[draw=#1,thick,inner sep=2pt]{test}{\myfontsize #2}}

\newcommand{\mystartPosBox}{      
    \mytextbox{myred}{
        ...\textbf{\textcolor{myred}{the}} Seattle Seahawks...
    }
}
\newcommand{\myendPosBox}{
    \mytextbox{myred}{
        ...the Seattle \textbf{\textcolor{myred}{Seahawks}}...
    }
}
\newcommand{\myPosBoxOne}{      
    \mytextbox{myred}{
        ...Seattle \textbf{\textcolor{myred}{Seahawks}} as...
    }
}
\newcommand{\myPosBoxTwo}{      
    \mytextbox{myred}{
        ...the \textbf{\textcolor{myred}{Seahawks}} became...
    }
}

\newcommand{\Ara}{ar}
\newcommand{\Cze}{cs}
\newcommand{\Gre}{el}
\newcommand{\Hin}{hi}
\newcommand{\Heb}{iw}
\newcommand{\Jap}{jp}
\newcommand{\Kor}{ko}
\newcommand{\Mal}{ml}
\newcommand{\Mon}{mn}
\newcommand{\Pol}{pl}
\newcommand{\Rus}{ru}
\newcommand{\Tal}{ta}
\newcommand{\Tha}{th}
\newcommand{\Tur}{tr}
\newcommand{\Chi}{zh}

\title{
    \Np\ Masked Language Modeling
}

\newcommand{\sewon}[1]{
    \textcolor{magenta}{[Sewon: #1]}
}

\newcommand{\affilsup}[1]{\rlap{\textsuperscript{\normalfont#1}}}
\author{
    Sewon Min\affilsup{1,2} \qquad
    Weijia Shi\affilsup{1,2} \qquad
    Mike Lewis\affilsup{2} \qquad
    Xilun Chen\affilsup{2}
    \\
    \textbf{Wen-tau Yih}\affilsup{2} \qquad
    \textbf{Hannaneh Hajishirzi}\affilsup{1,3} 
    \qquad
     \textbf{Luke Zettlemoyer}\affilsup{1,2} \\
    $^1$University of Washington \qquad
    $^2$Meta AI \qquad
    $^3$Allen Institute for AI \\
    \texttt{\{sewon,swj0419,hannaneh,lsz\}@cs.washington.edu} \\
    \texttt{\{mikelewis,xilun,scottyih\}@meta.com}
}

\begin{document}
\maketitle
\begin{abstract}
    Existing language models (LMs) predict tokens with a softmax over a finite vocabulary, which can make it difficult to predict rare tokens or phrases.
    We introduce \ours, the first \np\ masked language model that replaces this softmax with a \np\ distribution over every {\em \phrase} in a reference corpus.
    \ours\ fills in the \mask\ solely from retrieving a token from a text corpus.
    We show that \ours\ can be efficiently trained with a contrastive objective and an in-batch approximation to full corpus retrieval.
    Zero-shot evaluation on 16 tasks including classification, fact probing and question answering demonstrates that \ours\ outperforms significantly larger parametric models, either with or without a retrieve-and-generate approach.
    It is particularly better at dealing with rare patterns (word senses or facts) and predicting rare or nearly unseen words (e.g., non-Latin script). We release the model and code at \href{https://github.com/facebookresearch/NPM}{\texttt{github.com/facebookresearch/NPM}}.
\end{abstract}

\section{Introduction}\label{sec:intro}
\begin{figure}[t]
\centering \footnotesize
\resizebox{\columnwidth}{!}{\includegraphics{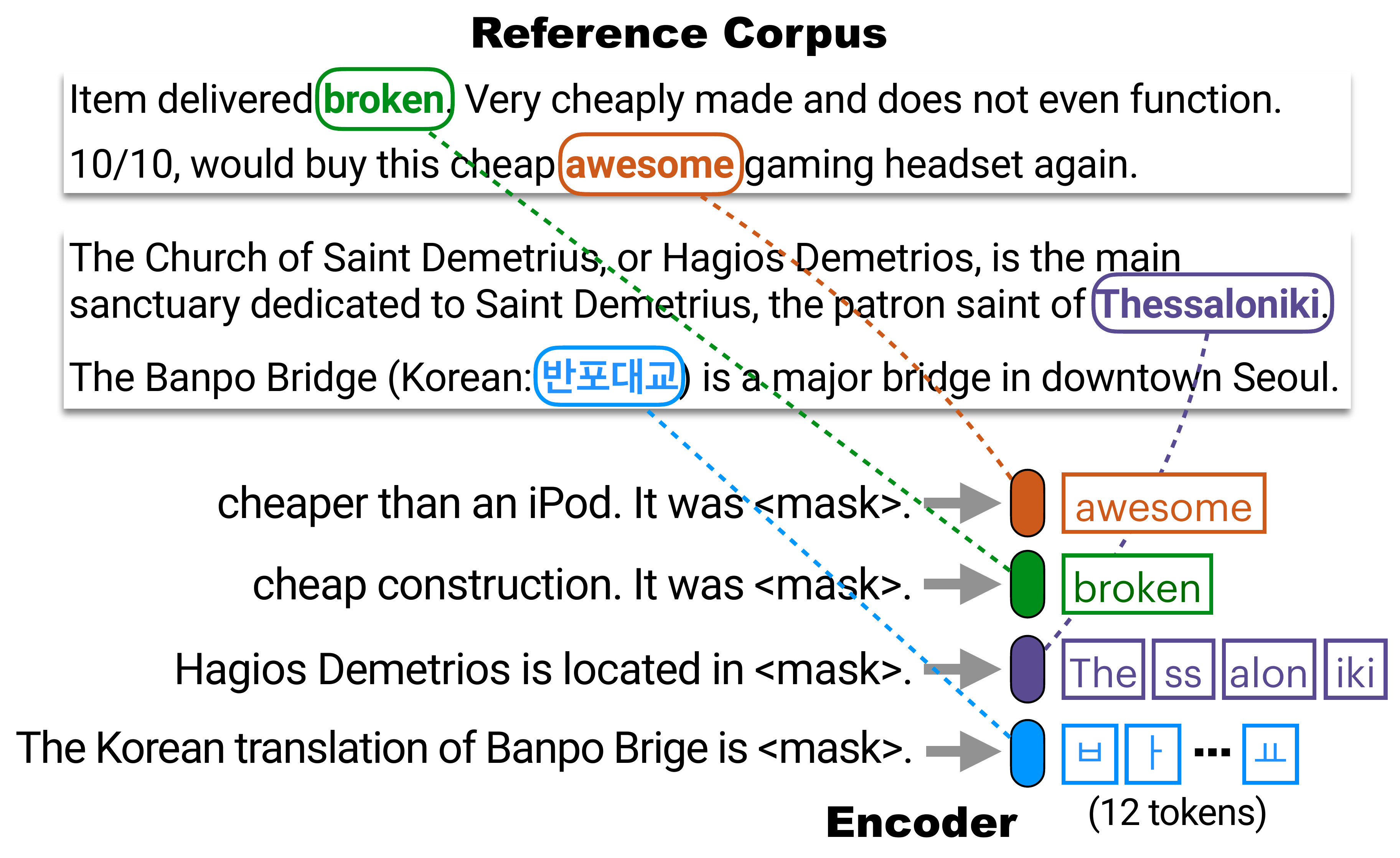}}\vspace{-.3em}
\caption{
An illustration of \ours. The {\em encoder} maps a masked sentence into a dense vector, and retrieves the nearest phrase from a {\em reference corpus}. \ours\ can fill in the \mask\ with multiple tokens, e.g., {\em Thessaloniki} (4 BPE tokens) and unseen words, e.g., \begin{CJK}{UTF8}{mj}반포대교\end{CJK}
(12 BPE tokens). 
}\label{fig:intro}
\end{figure}

Current large language models, despite their wide use and impressive performance, are expensive to scale, difficult to update, and struggle with long-tail knowledge and patterns~\citep{kandpal2022large}.
Recent work follows a retrieve-and-generate approach to partially address these issues~\citep{lewis2020retrieval,izacard2022few}; however, their final predictions are still made by a parametric model.
In particular, they still include a softmax over a finite vocabulary, which limits expressivity~\citep{yang2018breaking,pappas-etal-2020-grounded} and can make them reluctant to predict rare or unseen tokens (e.g., {\em Thessaloniki} in Figure~\ref{fig:intro}).
%


In this paper, we introduce \ours, the first  \textbf{N}on\textbf{\textsc{p}}arametric \textbf{M}asked Language Model that predicts tokens solely based on a \np\ distribution over {\em \phrases} in a text corpus (Figure~\ref{fig:intro}). 
\ours\ consists of an {\em encoder} that maps the text into a fixed-sized vector, and a {\em reference corpus} from which \ours\ retrieves a \phrase\ and fills in the \mask.
It, crucially, does not have a softmax over a fixed vocabulary, but instead has a {\em fully \np} distribution over phrases. This is in contrast to a recent body of work that incorporates \np\ components in a parametric model~\citep{borgeaud2022improving,izacard2022few,zhong2022training}.

Training such a \np\ model introduces two key challenges: (1) full corpus retrieval during training is expensive, and (2)
learning to predict an arbitrary length \phrase\ without a decoder is non-trivial.
We address the first challenge by using in-batch approximations to full corpus retrieval~\citep{wu2020scalable,zhong2022training}, and the second by extending span masking~\citep{joshi2020spanbert} and a phrase-level contrastive objective~\citep{oord2018representation,lee2021learning}. 

%

We perform zero-shot evaluation on 16 tasks including classification, fact probing and question answering. They include temporal shift and word-level translation tasks that highlight the need to predict new facts or rare phrases. 
We compare with a range of competitive baselines including encoder-only~\citep{liu2019roberta}, encoder-decoder~\citep{raffel2020exploring}, and decoder-only models~\citep{zhang2022opt,brown2020language}.
We also compare with a retrieve-and-generate approach that feeds a concatenation of the input and passages to parametric models using off-the-shelf retrieval.
%
%
%
%
%
%
Results show that \ours\ is significantly more parameter-efficient, outperforming up to 500x larger parametric models and up to 37x larger retrieve-and-generate models.
It is particularly good at (1) predicting rare words (e.g., an entity split into multiple BPE tokens such as {\em Thessaloniki}) and (2) disambiguating word senses (e.g., {\em cheap} may indicate {\em inexpensive} or {\em of very poor quality}; Figure~\ref{fig:intro}).
Finally, our evaluation on an entity translation task demonstrates that \ours\ can predict a word consisting of characters that are extremely rare if not unseen (e.g., non-Latin script; Figure~\ref{fig:intro}). 

In summary, our contributions are as follows.
\vspace{-.5em}
\begin{enumerate}[leftmargin=15pt]\itemsep -.3em
    \item We introduce \ours, the first \np\ masked language model that fills in the \mask\ solely from a phrase-level \np\ distribution over a 
    corpus.
    \item We introduce a novel training scheme to train \ours\ on unlabeled data. 
    We completely remove the softmax over the output vocabulary, enabling an effectively unbounded output space by predicting any $n$-gram.
    \item Zero-shot evaluation on 16 downstream tasks shows that \ours\ outperforms significantly larger parametric models, are better on rare patterns, scale well, can be efficiently updated at test time, and can predict extremely rare if not unseen tokens (e.g., words in non Latin script).
\end{enumerate}

\section{Related Work}\label{sec:related}\paragraph{Language Models (LMs).}
Large LMs trained on a vast amount of text are shown to perform a wide range of downstream tasks in a zero-shot manner by converting a task into a cloze format~\citep{radford2019language,brown2020language}.
This is possible because a variety of knowledge is encoded in the parameters of the models.
Recent work has scaled parametric LMs by adding more parameters~\citep{brown2020language,rae2021scaling,chowdhery2022palm} which can be very expensive in practice.
Moreover, such models struggle with predicting rare words or entities, and cannot be updated over time. 
%

There has been a recent body of work
that incorporates the \np\ component with a parametric LM.
We distinguish (1) work that concatenates retrieved text to the input and trains the model with a standard LM objective (\citet{borgeaud2022improving,izacard2022few}; so-called retrieve-and-generate approaches) from (2) work that retrieves tokens from a large text corpus to estimate a probability distribution that is interpolated with the output distribution from a standard LM (\citet{khandelwal2020generalization,yogatama2021adaptive,zhong2022training,lan2023copy}; so-called \knn\ models).
Our work is closely related to such a line of work 
and can be seen as an extreme version of the \knn\ approach with no interpolation.
However, our work is the first that models a {\em fully} \np\ distribution by entirely removing the softmax over a finite vocabulary.
This offers a range of new functionalities, such as modeling a distribution over \phrases, or predicting rare or unseen words. 

\vspace{-.2em}
\paragraph{Bottleneck in softmax.} Most if not all language models use a softmax function that gives a categorical probability distribution over a finite vocabulary. \citet{yang2018breaking} showed that this softmax is a low-rank approximation of a high-rank output space, making the model less expressive. \citet{pappas-etal-2020-grounded} discussed that a fixed output vocabulary makes language models resistant to adaptation to new domains and tasks.
We share the motivation with such prior work and propose to use a \np\ output space to address these issues.
Moreover, although not explicitly explored in this paper, our work that completely removes the softmax over the vocabulary can make training more efficient, especially when the vocabulary is large (e.g., multilingual models~\citep{conneau2020unsupervised}).


\vspace{-.2em}
\paragraph{\Np\ models.}
In \np\ models, the data distribution is not defined by a fixed set of parameters, but is rather a function of the available data~\citep{siegel1957nonparametric,hollander2013nonparametric}.
Having complexity that grows as the data grows, they are differentiated from parametric models whose complexity is bounded as a priori.
\citet{freeman2002example} noted that the term \np\ does not imply that they have no parameters,
but rather that the number and nature of the {\em effective} parameters are flexible and can depend on the data.

Recent work in NLP has explored \np\ inference without training~\citep{khandelwal2020generalization,he-etal-2021-efficient,xu2022capturing}, or trained the \np\ model on the labeled data for a specific downstream task~\citep{seo2018phrase,seo2019real,lee2021learning}. 
In contrast, our work trains a fully \np\ language model without the labeled data and performs a range of tasks zero-shot.


\section{Method}\label{sec:method}\begin{figure}[t]
\centering \footnotesize
\resizebox{\columnwidth}{!}{\includegraphics{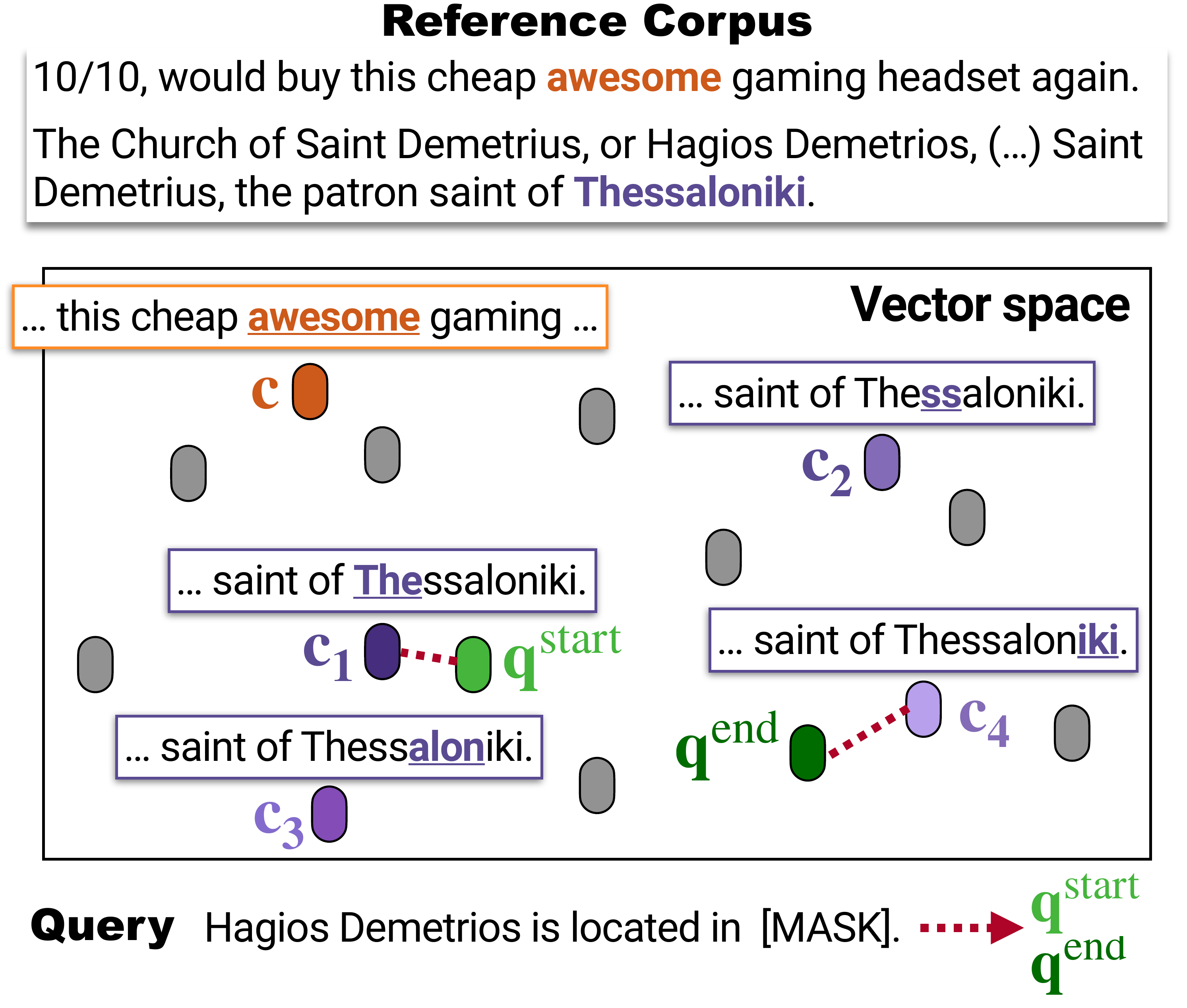}}\vspace{-.4em}
\caption{
Inference of \ours\ (Section~\ref{subsec:ours-inference}).
Each token in the reference corpus $\mathcal{C}$ is mapped into a dense vector space. 
At test time, a query is represented as two vectors, $\mathbf{q}^\mathrm{start}$ and $\mathbf{q}^\mathrm{end}$, each in the same vector space.
We use a nearest neighbor search to retrieve the start and the end of the \phrase\ using $\mathbf{q}^\mathrm{start}$ and $\mathbf{q}^\mathrm{end}$, respectively.
}\label{fig:inference-multi}
\end{figure}

We introduce \textbf{\ours}, the first \textbf{N}on\textbf{\textsc{p}}arametric \textbf{M}asked Language Model.
\ours\ consists of an encoder and a reference corpus, and models a \np\ distribution over a reference corpus (Figure~\ref{fig:intro}).
The key idea is to map all the phrases in the corpus into a dense vector space using the encoder and, when given a query with a \mask\ at inference, use the encoder to locate the nearest \phrase\ from the corpus and fill in the \mask.

Encoder-only models are competitive representation models~\citep{patel2022bidirectional}, outperforming the other two classes of models in classification tasks (Section~\ref{subsec:results-closed}). However, existing encoder-only models are unable to make a prediction whose number of tokens is unknown, making their use cases limited without fine-tuning. \ours\ addresses this issue, since it can fill in the \mask\ with an arbitrary number of tokens by retrieving a {\em \phrase}.

We first describe inference of \ours\ assuming a learned encoder (Section~\ref{subsec:ours-inference}), and then describe how we train the encoder to map the text into a good vector space (Section~\ref{subsec:ours-train}).

\subsection{\ours: Inference}\label{subsec:ours-inference}

\paragraph{Overview.}
The encoder maps every distinct {\em \phrase} in a reference corpus $\mathcal{C}$ into a dense vector space.
At test time, the encoder maps the masked query into the same vector space and retrieves \phrases\ from $\mathcal{C}$ to fill in the \mask. 
Here, $\mathcal{C}$ does not have to be the same as the training corpus, and can be replaced or scaled at test time without re-training the encoder.

In practice, there is a significant number of \phrases\ in the corpus, and it is expensive to index all of them.
%
We therefore use a technique from \citet{lee2021learning}
that represents a \phrase\ with {\em token} representations of the start and the end of the \phrase.
In this approach, we index representations of each distinct token in $\mathcal{C}$, and then at test time, use a $k$ nearest neighbor search for the start and the end of the phrase, separately.
Consider Figure~\ref{fig:inference-multi} as an example.
We represent a query with two vectors, \mygreen{$\mathbf{q}^\mathrm{start}$} and \mydarkgreen{$\mathbf{q}^\mathrm{end}$}. We then use each to retrieve the start and the end of the plausible \phrases---in this case, \myindigo{$\mathbf{c}_1$} and \myindigo{$\mathbf{c}_4$}, which are the start and the end of \myindigo{\textbf{\em Thessaloniki}}, respectively.

\begin{figure*}[t]
\centering \footnotesize
\resizebox{2.05\columnwidth}{!}{\includegraphics{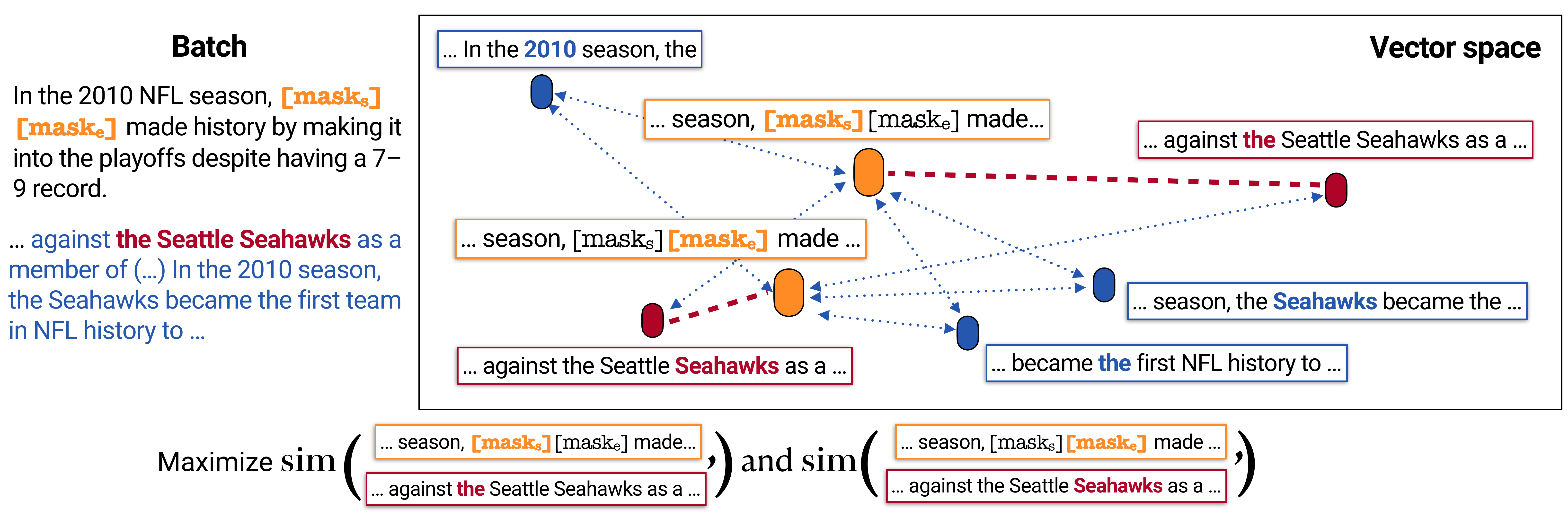}}\vspace{-.3em}
\caption{
Training of \ours. (Section~\ref{subsec:ours-train}).
\textcolor{orange}{\textbf{\maskS\maskE}} indicates the masked {\em span} whose original \phrase\ is {\em the Seattle Seahawks}. 
We maximize the similarity scores between \mytextbox{orange}{...\textbf{\textcolor{orange}{\maskS}}\maskE...} and\mystartPosBox, and between \mytextbox{orange}{... \maskS\textbf{\textcolor{orange}{\maskE}} ...} and \myendPosBox.
}\label{fig:training-multi}
\end{figure*}

\begin{figure}[t]
\centering \footnotesize
\resizebox{\columnwidth}{!}{\includegraphics{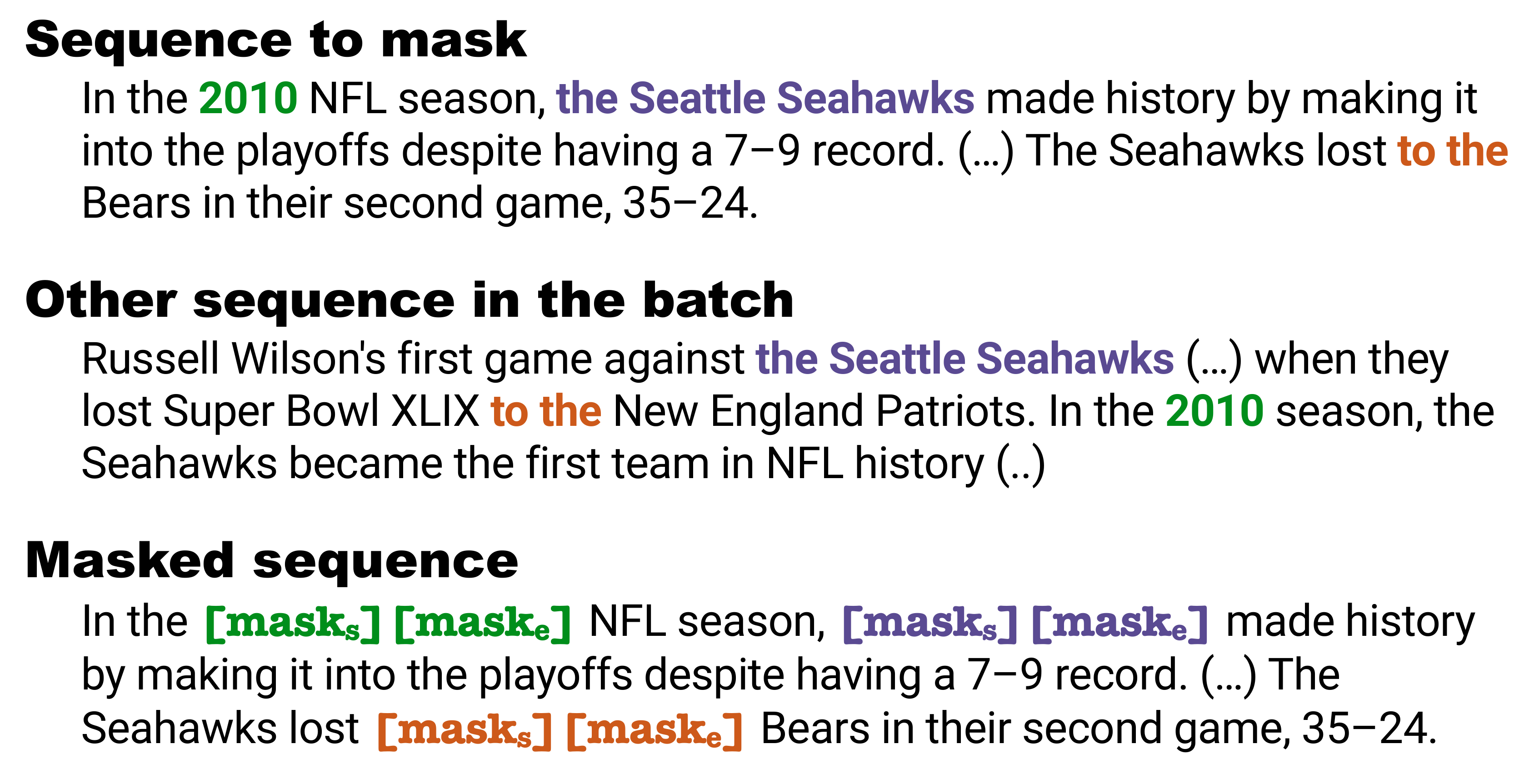}}\vspace{-.3em}
\caption{
Our span masking (Section~\ref{subsubsec:ours-train-masking}).
For simplicity, this figure assumes two sequences in the batch.
Spans to mask out are chosen based on whether there is any co-occurring spans in other sequences in the batch. Then, each span is replaced with \maskS\maskE.
}\label{fig:masking}
\end{figure}

\paragraph{Method.}
Formally, let $\mathcal{C} = \{c_1, \cdots, c_N \}$ be a reference corpus with $N$ tokens. 
We first map each token $c_i$ into a contextualized, $h$-dimensional vector $\mathbf{c}_i \in \mathbb{R}^h$ by feeding the text into the encoder and take the vector that corresponds to each token: 
$\mathbf{c}_1...\mathbf{c}_{N} = \mathrm{Encoder}(c_1...c_N).$


At inference time, \ours\ is given a query whose $t$-th token is masked: $q_1...q_{t-1},\mask,q_{t+1}...q_L$.
We replace \mask\ with two special tokens \maskS\maskE\ and feed it into the encoder to obtain a list of $h$-dimensional vectors:\begin{eqnarray*}
\mathbf{q}_1...\mathbf{q}_{L+1} = \mathrm{Encoder}(q_1...q_{t-1}, \maskS, \\ \maskE, q_{t+1}...q_L).\end{eqnarray*}
We then take the vector corresponding to \maskS\ and \maskE\ as $\mathbf{q}^\mathrm{start}$ and $\mathbf{q}^\mathrm{end}$, respectively.\footnote{This allows obtaining two vectors without encoding the query twice, e.g., unlike \citet{lee2021learning}}
$$\mathbf{q}^\mathrm{start}=\mathbf{q}_t,\mathbf{q}^\mathrm{end}=\mathbf{q}_{t+1}.$$
%
We then make a prediction via:
\begin{eqnarray*}
\argmax_{v^* \in \mathcal{V}^*} \sum_{i \leq j}\mathbb{I}[v^*=c_{i:j}]\bigg( ~~~~~~~~~~~~~~~~~~~~~~~~~~~~~~~~~~~~~~~~\\ \mathrm{exp}(\mathrm{sim}(\mathbf{q}^\mathrm{start},\mathbf{c}_i))+ \mathrm{exp}(\mathrm{sim}(\mathbf{q}^\mathrm{end},\mathbf{c}_j))\bigg),\end{eqnarray*} where $\mathcal{V}^*$ is a set of possible \ngrams\ defined by the vocabulary $\mathcal{V}$ and $\mathrm{sim}$ is a pre-defined similarity function that maps a pair of vectors into a scalar value. 
In practice, iterating over $N$ tokens is infeasible. We thus use an approximation using a fast nearest neighbor search for the start and the end separately. Details are provided in Appendix~\ref{app:model-details-ours}.

\newcommand{\myskip}[1]{}

\myskip{
We then consider a \phrase\ $c^*$, a list of consecutive tokens in $\mathcal{C}$ whose length can be arbitrary.
We represent $c^*$ as a $2h$-dimensional vector $\mathbf{c}^* = [\mathbf{c}^{s(c^*)}:\mathbf{c}^{e(c^*)}]$, where $[:]$ indicates a concatenation of two vectors, and $\mathbf{c}^{s(c^*)}$ and $\mathbf{c}^{e(c^*)}$ indicate the vectors corresponding to the start token and the end token of $c^*$, respectively.


At inference time, \ours\ is given a query with a \mask. The encoder first maps the query into two $h$-dimensional vectors: $\mathbf{q}^\mathrm{start}$ and $\mathbf{q}^\mathrm{end}$. 
We represent the query as $\mathbf{q} = [\mathbf{q}^\mathrm{start};\mathbf{q}^\mathrm{end}]$, and make a prediction via:$$\argmax_{v^* \in \mathcal{V}^*} \sum_{c^* \in \mathcal{C}^*}\mathbb{I}[v^*=c^*]\mathrm{exp}(\mathrm{sim}(\mathbf{q}, \mathbf{c}^*)),$$ where $\mathcal{V}^*$ is a set of possible \ngrams\ defined by the vocabulary $\mathcal{V}$, $\mathcal{C}^*$ is a set of \phrases\ in $\mathcal{C}$, and $\mathrm{sim}$ is a pre-defined similarity function that maps a pair of vectors into a scalar value. We assume $\mathrm{sim}([\mathbf{a};\mathbf{b}], [\mathbf{c};\mathbf{d}])=\mathrm{sim}(\mathbf{a},\mathbf{c})+\mathrm{sim}(\mathbf{b},\mathbf{d}),$ and use:
\begingroup\makeatletter\def\f@size{10}\check@mathfonts 
\begin{eqnarray*}
    & \mathrm{exp}(\mathrm{sim}(\mathbf{q},\mathbf{c}^*)) \\
    & = \mathrm{exp}(\mathrm{sim}([\mathbf{q}^\mathrm{start}:\mathbf{q}^\mathrm{end}],[\mathbf{c}^{s(c^*)}:\mathbf{c}^{e(c^*)}])) \\
    & = \mathrm{exp}(\mathrm{sim}(\mathbf{q}^\mathrm{start},\mathbf{c}^{s(c^*)}))+\mathrm{exp}(\mathrm{sim}(\mathbf{q}^\mathrm{end}, \mathbf{c}^{e(c^*)})).
\end{eqnarray*} \endgroup

\vspace{-.2em}
\paragraph{Approximation.}
In practice, iterating over all \phrases\ in $\mathcal{C}$ is infeasible. We thus use an approximation using a fast nearest neighbor search for the start and the end separately. We take the top $k$ {\em start} tokens and the top $k$ {\em end} tokens, and define an approximated set of candidate \phrases\ $\tilde{\mathcal{C}}^*$ as \phrases\ composed by these tokens (details provided in Appendix~\ref{app:model-details-ours}). We then predict:$$\argmax_{v^* \in \mathcal{V}^*} \sum_{c^* \in \tilde{\mathcal{C}}^*}\mathbb{I}[v^*=c^*]\mathrm{exp}(\mathrm{sim}(\mathbf{q}, \mathbf{c}^*)).$$
}

\vspace{-.2em}
\paragraph{Similarity function.}
The choice of similarity function can be flexible. We follow \citet{zhong2022training} in using a scaled inner product
$\mathrm{sim}(\mathbf{h}_1,\mathbf{h}_2)=\frac{\mathbf{h}_1 \cdot \mathbf{h}_2}{\sqrt{h}},$
where $h$ is a dimension of the token vectors.

\subsection{\ours: Training}\label{subsec:ours-train}

\ours\ is trained on unlabeled text data. 
We describe the masking strategy
first (Section~\ref{subsubsec:ours-train-masking}), and then the training objective (Section~\ref{subsubsec:ours-train-objective}).

\subsubsection{Masking}\label{subsubsec:ours-train-masking}
%
%
We extend span masking~\citep{joshi2020spanbert}, which masks spans (consecutive tokens) whose length is sampled from a geometric distribution.
Our span masking differs from \citet{joshi2020spanbert} in two ways.
First, we mask spans if they co-occur in the other sequences in the batch to guarantee in-batch positives during training (Section~\ref{subsubsec:ours-train-objective}).
For instance, masked spans in Figure~\ref{fig:masking} are `\mygreen{\em 2010}', `\myindigo{\em the Seattle Seahawks}' and `\myorange{\em to the}' all of which are found in the other sequences.
Second, instead of replacing each token in the span with a \mask, we replace the whole span with two special tokens \maskS\maskE.
For instance, each of  `\mygreen{\em 2010}', `\myindigo{\em the Seattle Seahawks}' and `\myorange{\em to the}' is replaced with \maskS\maskE. 
This is to obtain the start and the end vectors for each span as we do at inference. 

\subsubsection{Training Objective}\label{subsubsec:ours-train-objective}

\paragraph{Key idea.}
We illustrate an example in Figure~\ref{fig:training-multi}.
The masked span is `\textcolor{myred}{\em the Seattle Seahawks}', thus the model should retrieve a \phrase\ `\textcolor{myred}{\em the Seattle Seahawks}' from other sequences in the reference corpus when it is given a query like this at test time.
%
%
Specifically,
we should encourage the \maskS\ vector to be closer to \mystartPosBox\ and the \maskE\ vector to be closer to \myendPosBox, while being distant from other tokens.
We train the model to do so by approximating the full corpus as the other sequences in the batch.
Concretely, we train the model to retrieve the start and the end of the span `\textcolor{myred}{\em the Seattle Seahawks}' from other sequences in the same batch. Note that our masking strategy ensures that every masked span has a co-occurring span in the batch (Section~\ref{subsubsec:ours-train-masking}). 


\newcommand{\gxsimple}{g(x_j)}
\newcommand{\ysPossimple}{\mathcal{Y}_\mathrm{s}^+(\gxsimple)}
\newcommand{\ysNegsimple}{\mathcal{Y}_\mathrm{s}^-(\gxsimple)}
\newcommand{\yePossimple}{\mathcal{Y}_\mathrm{e}^+(\gxsimple)}
\newcommand{\yeNegsimple}{\mathcal{Y}_\mathrm{e}^-(\gxsimple)}

\newcommand{\gx}{g_t^i}
\newcommand{\ysPos}{\mathcal{Y}_\mathrm{s}^+(\gx)}
\newcommand{\ysNeg}{\mathcal{Y}_\mathrm{s}^-(\gx)}
\newcommand{\yePos}{\mathcal{Y}_\mathrm{e}^+(\gx)}
\newcommand{\yeNeg}{\mathcal{Y}_\mathrm{e}^-(\gx)}

\vspace{-.2em}
\paragraph{Obtaining vector representations.}
Consider the $i$-th sequence in the batch that consists of $L$ tokens, $x^i=x^i_1...x^i_L$.
We denote $\hat{x}^i=\hat{x}^i_1...\hat{x}^i_L$ as a consequence of span masking over $x^i$.
Both $x^i$ and $\hat{x}^i$
are fed into the encoder, and each token is mapped into an $h$-dimensional vector:\footnote{The unmasked sequence and the masked sequence may have different lengths before padding, but we pad them to have the same length.}
\begin{eqnarray*}
    \mathbf{x}_1^i \cdots \mathbf{x}_L^i&=&\mathrm{Encoder}(x_1^i \cdots x_L^i), \\
    \mathbf{\hat{x}}_1^i \cdots \mathbf{\hat{x}}_L^i&=&\mathrm{Encoder}(\hat{x}_1^i \cdots \hat{x}_L^i).
\end{eqnarray*}

\myskip{
\begin{figure*}[t]
\centering \footnotesize
\resizebox{2\columnwidth}{!}{\includegraphics{imgs/npm-training.png}}\vspace{-.3em}
\caption{
The high-level idea of training \textbf{\ourssingle} (Section~\ref{subsec:ours-single}).
\textcolor{orange}{\textbf{\mask}} indicates the masked token whose original token is {\em Seahawks}. %
\textcolor{myred}{Positives} are tokens from other sequences in-batch that share the vocab: \myPosBoxOne\ and \myPosBoxTwo.
\textcolor{myblue}{Negatives} are tokens from other sequences in-batch that are not positives (in the \textcolor{myblue}{blue} box).
We maximize the similarity scores between the masked token and positives, and minimize the similarity scores between the masked token and negatives. \sewon{Could just remove this figure if it feels repetitive.}
}\label{fig:training}
\end{figure*}
}

\vspace{-.2em}
\paragraph{Training objective.}
We consider a masked span in $x_i$, represented with \maskS\maskE, denoted as $\hat{x}_t^i, \hat{x}_{t+1}^i$.
We then denote $\gx$ as the original \ngram\ that were replaced by $\hat{x}_t^i, \hat{x}_{t+1}^i$.

We now define the objective for this masked span, and the final objective is summed over all masked spans. 
The training objective for this masked span is defined as\begingroup\makeatletter\def\f@size{10}\check@mathfonts
\begin{eqnarray*}
    - \Bigg( \mathrm{log}\frac{
        \sum_{\mathbf{y} \in \ysPos} \mathrm{exp}(\mathrm{sim}(\mathbf{\hat{x}}^i_{t}, \mathbf{y}))
    }{
        \sum_{\mathbf{y} \in \ysPos \cup \ysNeg} \mathrm{exp}(\mathrm{sim}(\mathbf{\hat{x}}^i_{t}, \mathbf{y}))
    }~~~~~~~~~~~~~~~~~\\
    ~~~~~~~~+\mathrm{log}\frac{
        \sum_{\mathbf{y} \in \yePos} \mathrm{exp}(\mathrm{sim}(\mathbf{\hat{x}}^i_{t+1}, \mathbf{y}))
    }{
        \sum_{\mathbf{y} \in \yePos \cup \yeNeg} \mathrm{exp}(\mathrm{sim}(\mathbf{\hat{x}}^i_{t+1}, \mathbf{y}))
    } \Bigg).
\end{eqnarray*} \endgroup
Here,
$\mathrm{sim}(\cdot,\cdot)$ is a similarity function defined in Section~\ref{subsec:ours-inference},
and $\ysPos$, $\ysNeg$, $\yePos$ and $\yeNeg$ are {\em start positives}, {\em start negatives}, {\em end positives} and {\em end negatives} of $\gx$, respectively, which are defined in the next paragraph.
%
This objective follows a phrase-level contrastive learning objectives in prior work~\citep{lee2021learning,ram-etal-2021-shot,deng-etal-2021-reasonbert,kulkarni-etal-2022-learning}
with an extension that allows {\em multiple} positives.

\vspace{-.15em}
\paragraph{In-batch positives and negatives.}
%
The start positives and the end positives are the start and the end of the spans to be retrieved.
The start negatives and the end negatives are tokens that are not the start positives and not the end positives, respectively.
%
%
\myskip{\vspace{-.1em}
\begin{itemize}[leftmargin=15pt]\itemsep -.15em
    \item a {start positive} of $\gx$ {\em if} 
    a span of length $|\gx|$ \textbf{starting} with $x^j_m$ is identical to $\gx$. 
    \item a {start negative} of $\gx$ {\em if} 
    a span of length $|\gx|$ \textbf{starting} with $x^j_m$ is \textbf{not} identical to $\gx$. 
    \item an {end positive} of $\gx$ {\em if} 
    a span of length $|\gx|$ \textbf{ending} with $x^j_m$ is identical to $\gx$. 
    \item an {end negative} of $\gx$ {\em if} 
    a span of length $|\gx|$ \textbf{ending} with $x^j_m$ is \textbf{not} identical to $\gx$. 
\end{itemize}}
More formally:\begingroup\makeatletter\def\f@size{10}\check@mathfonts
\begin{eqnarray*}
    \ysPos &=& \big\{ x^{j}_m | \gx=x^{j}_m...x^{j}_{m+|\gx|-1}~~\&~~i \neq {j}\big\}, \\
    \ysNeg &=& \big\{ x^{j}_m | \gx \neq x^{j}_m...x^{j}_{m+|\gx|-1}~~\&~~i \neq {j}\big\}, \\
    \yePos &=& \big\{ x^{j}_m | \gx=x^{j}_{m-|\gx|+1}...x^{j}_m~~\&~~i \neq {j}\big\}, \\
    \yeNeg &=& \big\{ x^{j}_m | \gx \neq x^{j}_{m-|\gx|+1}...x^{j}_m~~\&~~i \neq {j}\big\}.
\end{eqnarray*} \endgroup
Here, $|\gx|$ indicates the length of the span $\gx$.

\myskip{
\subsection{A special case: \ourssingle}\label{subsec:ours-single}
Along with \ours, we introduce \textbf{\ourssingle}, which outputs a \np\ distribution over every single {\em token} in $\mathcal{C}$, instead of a {\em \phrase}.
To some extent, \ours\ is a strict generalization of \ourssingle, and \ourssingle\ still can only fill in the \mask\ with a single token.
We however think \ourssingle\
can be useful for some applications, e.g., when it is used for fine-tuning, as existing encoder-only models are used for. 


We provide an overview here; see Appendix~\ref{app:model-details-ours} for details. 

\vspace{-.3em}
\paragraph{Inference.} As in \ours, each distinct token is represented as a $h$-dimensional vector.
However, different from \ours\ that represents a query with a $2h$-dimensional vector using \maskS\maskE, \ourssingle\ obtains a $h$-dimensional vector directly from a \mask. It then retrieves the $k$ nearest tokens from $\mathcal{C}$ and aggregate scores.

\vspace{-.3em}
\paragraph{Masking for training.}
We choose spans to mask out in the same way as in \ours. Then, instead of replacing the whole span with \maskS\maskE, we replace each token in the span with a \mask. This is closer to previous span masking~\citep{joshi2020spanbert}, except we still ensure the masked spans co-occur in the other sequences in the batch.

\vspace{-.3em}
\paragraph{Training objective.} We use a contrastive objective that encourages the vector corresponding to each \mask\ is close to the vectors corresponding to {\em positives}, and distant from the vectors corresponding to {\em negatives}. Here, positives and negatives are tokens from other sequences in the batch that correspond and do not correspond to the masked token, respectively.
For instance, if there is a masked token that corresponds to `{\em Seahawks}', positives can include \myPosBoxOne\ and \myPosBoxTwo. 


}

\section{Training Details}\label{sec:setup}\paragraph{Training data.}
We use English Wikipedia (August 2019) and an English portion of CC-News (\citet{mackenzie2020cc}, February 2019) for training, which contains 13B tokens in total. The data is segmented into sequences, each with up to 256 tokens.

\vspace{-.2em}
\paragraph{Training.}
We use the model architecture and initial weights of RoBERTa large~\citep{liu2019roberta}, consisting of 354M parameters.
Training is done for 100,000 steps, using thirty-two 32GB GPUs.
One batch consists of 512 sequences (131,072 tokens). 
We use an Adam optimizer~\citep{kingma2014adam} with a learning rate of $3 \times 10^{-5}$, weight decay of $0.01$ and $4,000$ steps of warm-up.

\vspace{-.2em}
\paragraph{Batching.}
The choice of batching is important in in-batch approximations, as it determines the quality of positives and negatives.
For instance, \citet{zhong2022training} uses BM25 to ensure the sequences in the same batch are likely to share the same topic.
With a pretraining corpus with billions of tokens, it can be significantly expensive to build a BM25 index. 
Therefore, we instead construct the batch by grouping sequences from the same document and assigning them to the same batch.\footnote{Documents that are not long enough to construct a batch are grouped with each other.} This trick ensures that (a) positives (spans that share the string) are likely to share the context, reducing false positives, and (b) negatives are those that the model is likely to be confused with, thus training against them helps the model better identify positives.
During training, we gather all sequences from multiple GPUs to increase the size of the effective batch and make in-batch approximation more effective.
\section{Experiments: Closed-set Tasks}\label{sec:exp-closed}
\begin{table*}[t]
    \centering \myfontsize 
    \setlength{\tabcolsep}{3.5pt}
    \begin{tabular}{l @{\hspace{-1em}} r rrrrrrrrra }
        \toprule
            Model & \# Params & 
            AGN &
            Yahoo &
            Subj &
            SST-2 &
            MR &
            RT &
            CR & 
            Amz &
            RTE &
            Avg
            \\
        \midrule
            \multicolumn{11}{l}{\textbf{\em Baselines (encoder-only)}} \\
            RoBERTa~\citep{gao2021making} & 1.0x & - & - & 51.4 & 83.6 & 80.8 & - & 79.5 & - & 51.3 & - \\
            RoBERTa & 1.0x  & 71.3 & 41.4 & 67.6 & 84.5 & 81.7 & 81.1 & 80.4 & 83.5 & 57.4 & 72.1 \\
        \midrule
            \multicolumn{11}{l}{\textbf{\em Baselines (encoder-decoder)}} \\
            T5        & 2.2x    & 72.0 & 51.3 & 54.9 & 57.5 & 57.7 & 59.1 & 56.4 & 59.3 & 55.6 & 58.2 \\
            T5 3B     & 8.5x    & \textbf{80.5} & 53.6 & 54.8 & 59.6 & 58.6 & 57.3 & 53.7 & 57.0 & 58.5 & 59.3 \\
        \midrule
            \multicolumn{11}{l}{\textbf{\em Baselines (decoder-only)}} \\
            GPT-2~\citep{shi2022nearest}   & 2.2x &
                67.4 & 49.7	& 60.8 & 55.3	& 54.6 & 53.0 & 66.2 & 57.6 & 53.1 & 57.5 \\
            ~~~~+ PMI~\citep{shi2022nearest}   & 2.2x &
                65.1 & 48.8	& 62.5 & 76.5	& 74.6 & 74.1 & 82.8 & 76.2 & 54.2 & 68.3  \\
            GPT-2 \knn$^\dagger$~\citep{shi2022nearest}      & 2.2x &
                29.8 & 37.0	& 50.0 & 47.1	& 49.9 & 49.1 & 69.3 & 57.4 & 54.1 & 49.3  \\
            GPT-2 \knnlm$^\dagger$~\citep{shi2022nearest}      & 2.2x &
                78.8 & 51.0	& 62.5 & 84.2	& 78.2 & 80.6 & 84.3 & 85.7 & 55.6 & 73.4 \\
            GPT-3~\citep{holtzman2021surface} & 500x        & 75.4 & 53.1 & 66.4 & 63.6 & 57.4 & 57.0 & 53.8 & 59.4 & 56.0 & 60.2 \\
            ~~~~+ PMI~\citep{holtzman2021surface} & 500x    & 74.7 & 54.7 & 64.0 & 71.4 & 76.3 & 75.5 & 70.0 & 75.0 & \textbf{64.3} & 69.5 \\
        \midrule
            \multicolumn{11}{l}{\textbf{\em Ours (encoder-only, \np)}} \\
            \ours$^\dagger$      & 1.0x &
            74.5 & 53.9 & \textbf{75.5} & \textbf{87.2} & \textbf{83.7} & \textbf{86.0} & 81.2 & 83.4 & 61.7 & \textbf{76.4} \\
        \midrule
            \multicolumn{11}{l}{\textbf{\em Full fine-tuning (reference)}} \\
            RoBERTa~\citep{gao2021making}           & 1.0x & - & - &  97.0 & 95.0 & 90.8 & - &  89.4 & - & 80.9 & - \\
        \bottomrule
    \end{tabular}\vspace{-.1em}
    \caption{
        Zero-shot results on closed-set tasks. {\em \# Params} indicates the relative number of model parameters compared to RoBERTa large (354M). RoBERTa, T5 and GPT-2 are their {\em large} variants unless specified otherwise;
        GPT-3 is from {\em Davinci, non-instruct}.
        Numbers with citations are taken from the corresponding papers. 
        As a reference, we provide results of fine-tuning on the full training dataset in the last row.
        $^\dagger$ indicates a reference corpus is used.
        \textbf{\ours\ significantly outperforms larger parameters models.}
    }\label{tab:closed-set}
\end{table*}

We perform zero-shot evaluation on closed-set tasks where a small set of candidates is given. 



\subsection{Evaluation Datasets}\label{subsec:closed-datasets}

We include nine classification datasets that are known for not necessarily requiring factual knowledge: AGNews~\citep{zhang2015character}, Yahoo~\citep{zhang2015character}, Subj~\citep{pang2004sentimental}, SST-2~\citep{socher2013recursive}, MR~\citep{pang2004sentimental}, Rotten Tomatoes (RT), CR~\citep{hu2004mining}, Amazon polarity (Amz, \citet{mcauley2013hidden}) and RTE~\citep{dagan2005pascal}.
The tasks range from topic classification and sentiment analysis to subjectivity classification and textual entailment.
Statistics are provided in Appendix~\ref{app:eval-details}.

\subsection{Baselines}\label{subsec:closed-baselines}
We compare 
with the encoder-only, the decoder-only and the encoder-decoder models with various sizes (354M to 175B parameters).
We include RoBERTa~\citep{liu2019roberta} as the encoder-only, T5~\citep{raffel2020exploring} as the encoder-decoder, and GPT-2/3~\citep{radford2019language,brown2020language} as the decoder-only model.
%
For the decoder-only models, we additionally apply PMI~\citep{holtzman2021surface} for better calibration of the model output.
We also compare with \citet{shi2022nearest} who use \knn\ inference using GPT-2 with PMI. In particular,
(1) GPT-2 \knn\ uses \knn\ inference without training, and
(2) GPT-2 \knnlm\ interpolates distributions from GPT-2 and GPT-2 \knn.

\subsection{Setup}\label{subsec:closed-setup}

We use the templates and verbalizers from \citet{shi2022nearest} for all models.
When available, we use fuzzy verbalizers from \citet{shi2022nearest}. 
%
We use a domain-specific reference corpus: a union of the English Wikipedia and CC News for AGN, Yahoo and RTE, a subjectivity corpus for Subj, and a review corpus for sentiment classification datasets. Their sizes vary from 15M tokens to 126M tokens. Details are in Appendix~\ref{app:eval-details}.
Fast similarity search is done using FAISS~\citep{johnson2019billion} with the HNSW index.
We use $k=4096$ for inference. 

\subsection{Results}\label{subsec:results-closed}
\ours~outperforms baselines in the zero-shot setting (Table~\ref{tab:closed-set}). We discuss the results in detail below.

\vspace{-.2em}
\paragraph{Comparison between baselines.}
Among parametric models, RoBERTa achieves the best performance, outperforming larger models including GPT-3.
This is perhaps surprising, and is likely because bidirectionality of the encoder-only model plays a vital role, as claimed in \citet{patel2022bidirectional}.
The \knnlm\ approach from \citet{shi2022nearest}, which incorporates the nonparametric component to the parametric model, outperforms all other baselines. Nonetheless, solely relying on retrieval (\knn) performs poorly with GPT-2, suggesting that using \knn\ at inference only is limited.

\begin{figure}[t]
\centering \footnotesize
\resizebox{0.9\columnwidth}{!}{\includegraphics{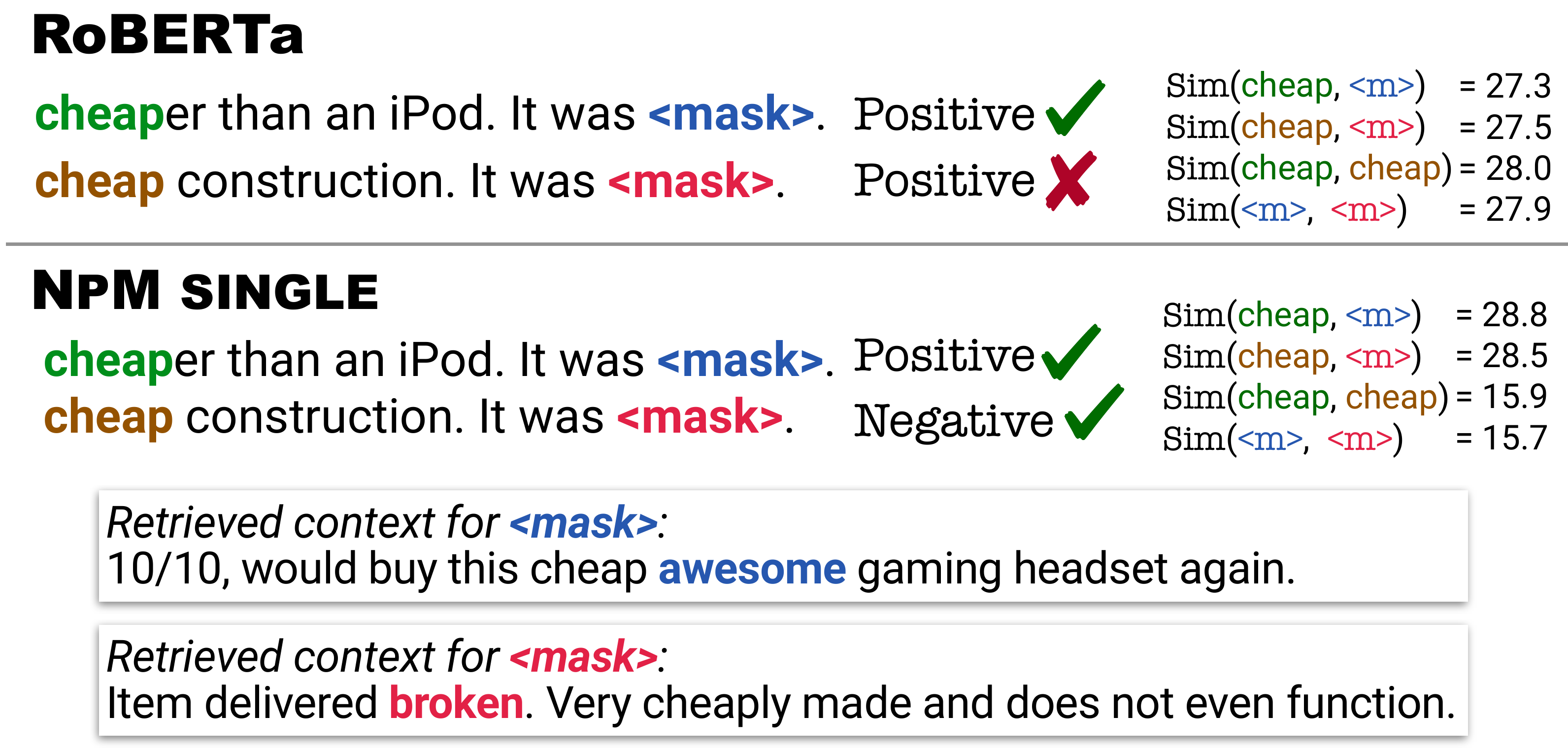}}\vspace{-.3em}
\caption{
Predictions from RoBERTa (baseline) and \ours.
The bottom indicates the context \ours\ retrieves to fill in \mask.
Note that the fuzzy verbalizer maps {\em broken} to \texttt{Negative} and {\em awesome} to \texttt{Positive}.
%
}\label{fig:prediction}
\end{figure}

\begin{figure*}[t]
\centering \footnotesize
\resizebox{2.05\columnwidth}{!}{
    \includegraphics[trim={4cm 2.5cm 4.5cm 4.3cm},clip]{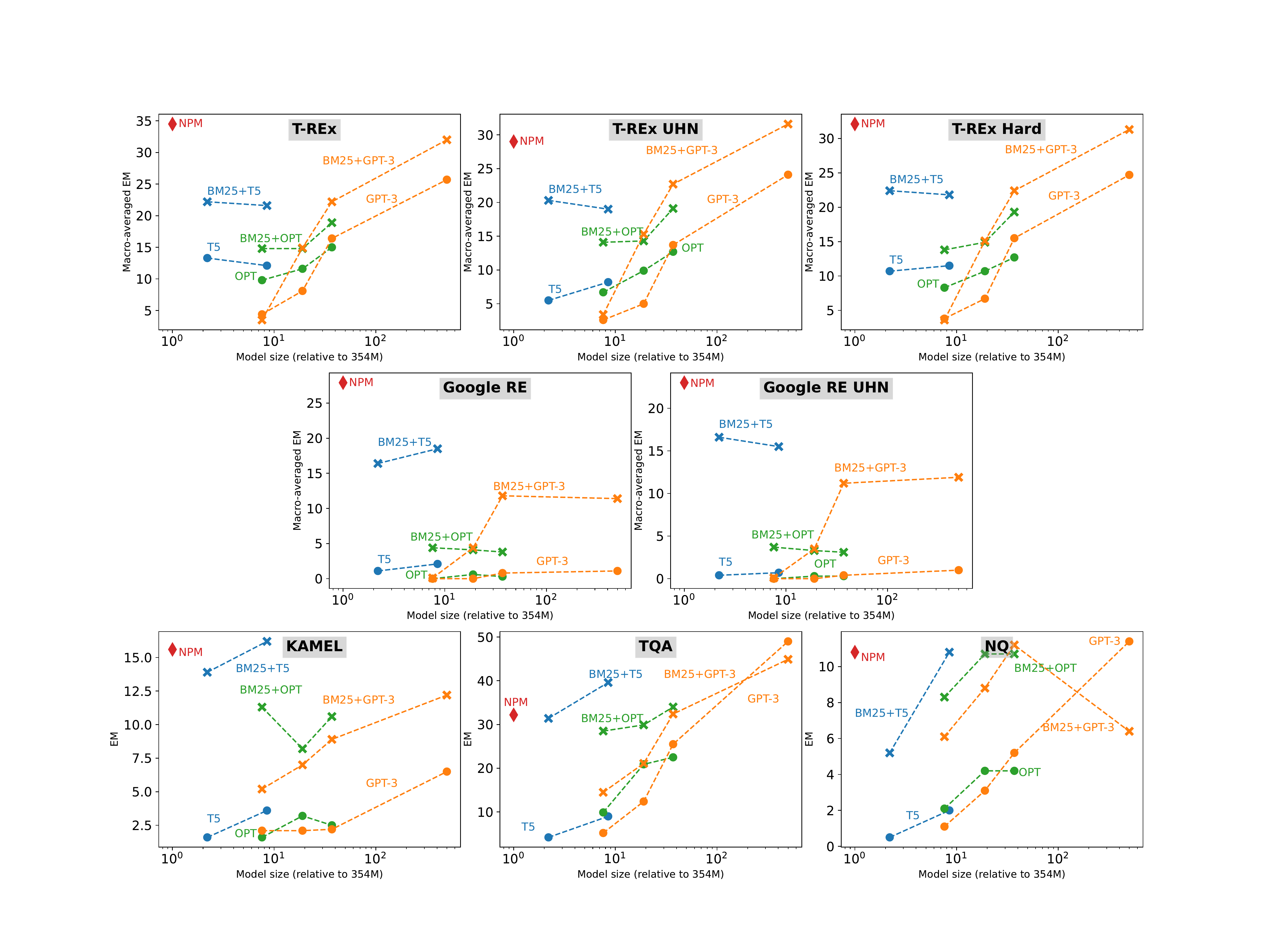}
}\vspace{-.5em}
\caption{
    \textbf{Zero-shot results on knowledge tasks.}
    The $x$-axis indicates the relative number of model parameters in log scale compared to RoBERTa large (354M).
    \ours\ outperforms significantly larger parameters models, either with or without BM25.
    See Table~\ref{tab:open-set} in Appendix~\ref{app:additional-results} for the raw numbers.
}\label{fig:open-set-results}
\end{figure*}

\vspace{-.2em}
\paragraph{Baselines versus \ours.}
\ours\ significantly outperforms all baselines,
achieving consistently competitive performance over all datasets.
This indicates that, even for tasks that do not explicitly require external knowledge, nonparametric models are very competitive.

\vspace{-.2em}
\paragraph{Qualitative analysis.}
Figure~\ref{fig:prediction} depicts predictions from RoBERTa and \ours\ on a sentiment analysis task. The first example uses \mygreen{\em cheap} to indicate {\em inexpensive}, and the second example uses \mybrown{\em cheap} to indicate {\em of very poor quality}.
RoBERTa predicts \texttt{Positive} to both, while \ours\ makes correct predictions by retrieving the context that uses {\em cheap} in the same context as the input.

We also find that representations from \ours\ lead to better word sense disambiguation.
For instance, RoBERTa assigns a high similarity score between \mygreen{\em cheap} ({\em inexpensive}) and \mybrown{\em cheap} ({\em of very poor quality}).
On the other hand, \ours\ successfully assigns a low similarity score between \mygreen{\em cheap} and \mybrown{\em cheap}, even though their surface forms are the same.

\section{Experiments: Open-set Tasks}\label{sec:exp-open}We include zero-shot evaluation on open-set tasks whose answer can be any arbitrary-length string.

\subsection{Evaluation Datasets}\label{subsec:open-datasets}

We evaluate on seven datasets: T-REx and Google-RE from LAMA~\citep{petroni-etal-2019-language}, KAMEL~\citep{kalo2022kamel}, Natural Questions (NQ, \citet{kwiatkowski2019natural}), TriviaQA (TQA, ~\citet{joshi-etal-2017-triviaqa}), \templama\ and an entity translation task.
In particular, TempLAMA requires probing knowledge with temporal updates, motivated by \citet{dhingra2022time} and \citet{jang2022temporalwiki}.
The entity translation task involves a translation of an entity from English to other, non-Latin languages, requiring the model to predict extremely rare (if not unseen) characters.
See Appendix~\ref{app:eval-details} for details and statistics of all datasets.

\myskip{
We provide an overview of our evaluation datasets. See Appendix~\ref{app:eval-details} for details and statistics.

\vspace{-.2em}
\paragraph{Knowledge tasks.}
We include T-REx and Google-RE from LAMA~\citep{petroni-etal-2019-language}, KAMEL~\citep{kalo2022kamel}, Natural Questions (NQ, \citet{kwiatkowski2019natural}) and TriviaQA (TQA, ~\citet{joshi-etal-2017-triviaqa}).
When available, we additionally evaluate on
the UHN (UnHelpfulNames) subset from \citet{Poerner2019BERTIN} or the hard subset from \citet{zhong2021factual}.

\vspace{-.2em}
\paragraph{Temporal knowledge.}
Motivated by \citet{dhingra2022time} and \citet{jang2022temporalwiki}, we create \templama, which requires probing knowledge with temporal updates.
It consists of the {\em changed} set whose answers are different between 2019 and 2022 or the valid answer exists in 2022 but not in 2019, and the {\em unchanged} set whose answers are the same between 2019 and 2022.


\vspace{-.2em}
\paragraph{Entity Translation.}
In order to evaluate the ability of the models in predicting a word with extremely rare (if not unseen) characters, we create a benchmark that requires a translation of an entity from English to other, non-Latin languages.
Since our model and baselines are monolingual models that are primarily trained in English,\footnote{
    In practice, the English data contains substantial amount of non-English text, which leads to cross-lingual transferability~\citep{blevins2022language}.
} this benchmark requires them to predict nearly unseen characters.
We use 15 languages:
Arabic (\Ara), Czech (\Cze), Greek (\Gre), Hindi (\Hin), Hebrew (\Heb), Japanese (\Jap), Korean (\Kor), Malayalam (\Mal), Mongolian (\Mon), Polish (\Pol), Russian (\Rus), Tamil (\Tal), Thai (\Tha), Turkish (\Tur), and Chinese (\Chi).
Details of the data construction are provided in Appendix~\ref{app:eval-details}.

Entity translation is a vital and challenging task in real applications such as machine translation~\citep{Babych2003ImprovingMT,yan2018impact} and cross-lingual question answering~\citep{tydiqa,xorqa}.
It is often beyond a series of simple translations of each word, or spelling out its pronunciation~\citep{moore2003learning,hassan2007improving,Sun2017CrossLingualEA}. For instance, the Korean translation of {\em Banpo Bridge} in Figure~\ref{fig:intro} (\begin{CJK}{UTF8}{mj}반포대교\end{CJK}) is not the concatenation of the translations of {\em Banpo} and {\em Bridge} (\begin{CJK}{UTF8}{mj}반포 다리\end{CJK}). 
}

\subsection{Baselines}\label{subsec:open-baselines}
We compare with T5~\citep{raffel2020exploring} as the encoder-decoder, and GPT-3~\citep{brown2020language} and OPT~\citep{zhang2022opt} as the decoder-only models.
The encoder-only models 
are not applicable for open-set tasks since the number of tokens to predict is unknown.


Prior work found that a ``retrieve-and-generate'' approach that concatenates the input and passages from an off-the-shelf retrieval system is often helpful in knowledge-dependent tasks~\citep{kandpal2022large}. We add them as baselines, using up to five passages from BM25~\citep{robertson2009probabilistic}.

\subsection{Setup}\label{subsec:open-setup}

For all datasets, we report Exact Match (EM). 
The LAMA test data is biased toward frequent entities because they are filtered to only include answers that are single tokens based on BERT~\citep{devlin2019bert}.
Since we do not want our evaluation to be biased toward overly frequent entities,
we report a micro-averaged accuracy over the data whose answers are 1, 2, 3 and 4+ grams, respectively.
Other datasets do not have such filtering, therefore we report average EM.

As a reference corpus, we use the English Wikipedia from 08/01/2019, consisting of 810M tokens.
For \templama, we use the English Wikipedia from 08/01/2022, consisting of 858M tokens.


For \ours, we find combining with sparse retrieval significantly helps, likely because dense retrieval and sparse retrieval capture complementary features~\citep{karpukhin2020dense,seo2019real}.
In particular, we reduce the search space to the top 3 passages based on BM25 and perform dense search as done in \citet{kassner2020bert}.


\begin{figure}[t]
\centering \footnotesize
\resizebox{\columnwidth}{!}{\includegraphics{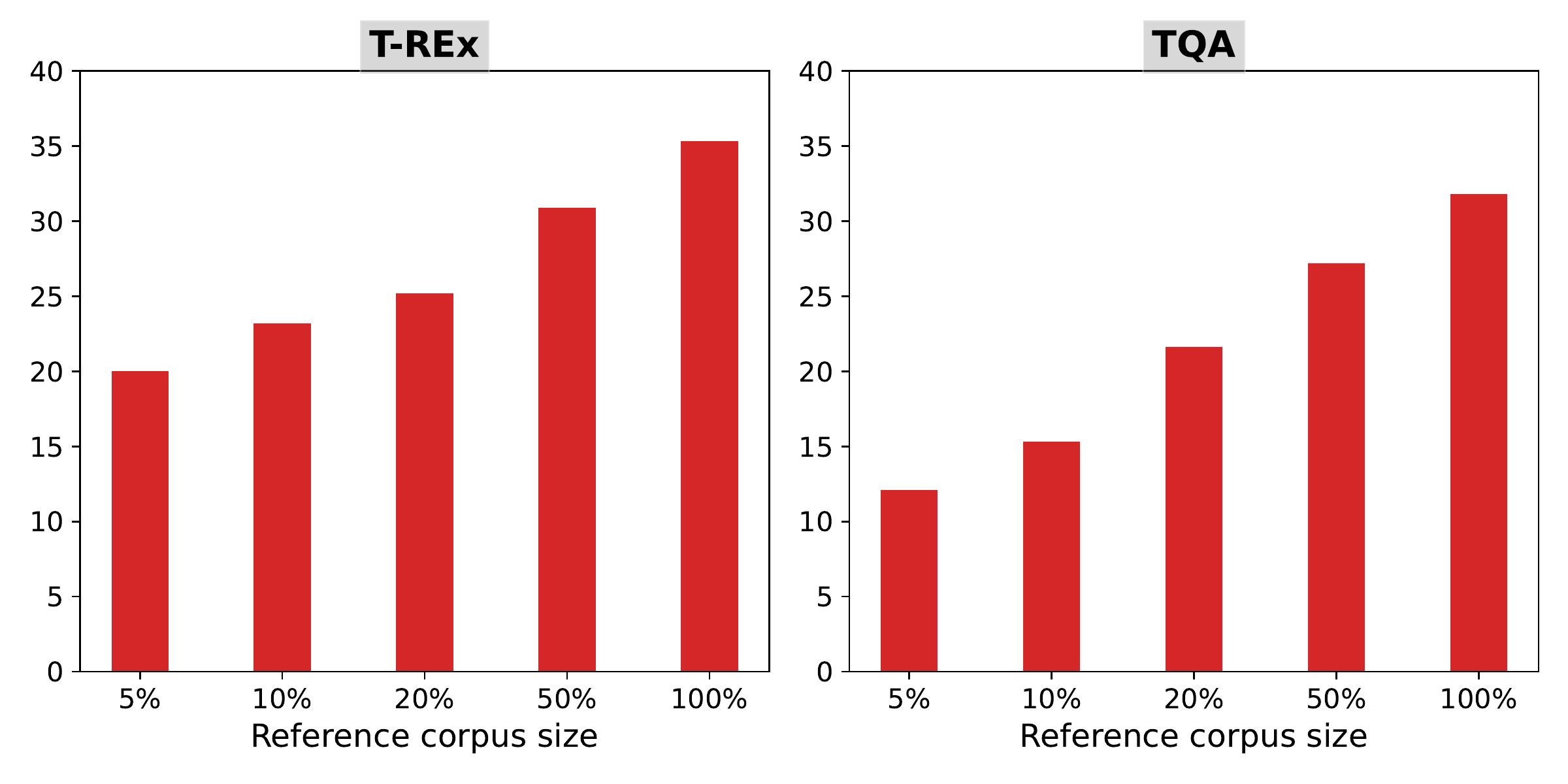}}\vspace{-.3em}
\caption{
    Ablation on the size of the reference corpus, from 41M tokens (5\%) to 810M tokens (100\%). There is a strong correlation between the size of the corpus and downstream performance.
}\label{fig:abl-corpus-size}
\end{figure}

\subsection{Results}\label{subsec:results-open}
Figure~\ref{fig:open-set-results} show results on five knowledge tasks. 

First, performance of parametric models largely depends on the number of parameters, as it has been claimed in much of prior work~\citep{brown2020language,kandpal2022large}.
The retrieve-and-generate approach
that combines parametric models with BM25 significantly improves performance.

\ours\ outperforms or is on par with significantly larger baselines across all datasets.
It substantially outperforms all models on two LAMA datasets, including 500x larger GPT-3 either with or without BM25.
On KML, TQA and NQ, \ours\ consistently outperforms 37x larger models with or without BM25.
This is impressive given that \ours\ is not trained on data with questions.

It is also worth noting that sparse retrieval is critical in \ours, e.g., without sparse retrieval, performance on LAMA-TREx drops from
34.5 to 16.1.
We think this is because (1) sparse retrieval and dense retrieval capture complementary features, and (2) the removal of approximation in search improves search quality.
We think future work can explore completely removing sparse retrieval, as has been done in \citet{lee2021learning} to improve \citet{seo2019real}.

\vspace{-.2em}
\paragraph{Impact of the reference corpus size.}
Figure~\ref{fig:abl-corpus-size} reports the impact of the size of the reference corpus, from 41M tokens (5\%) to 810M tokens (100\%).  Performance of \ours\ is highly correlated with the size of the reference corpus, strongly suggesting that using a larger reference corpus is important.

\begin{figure*}[t]
\centering \footnotesize
    \resizebox{2.1\columnwidth}{!}{\includegraphics{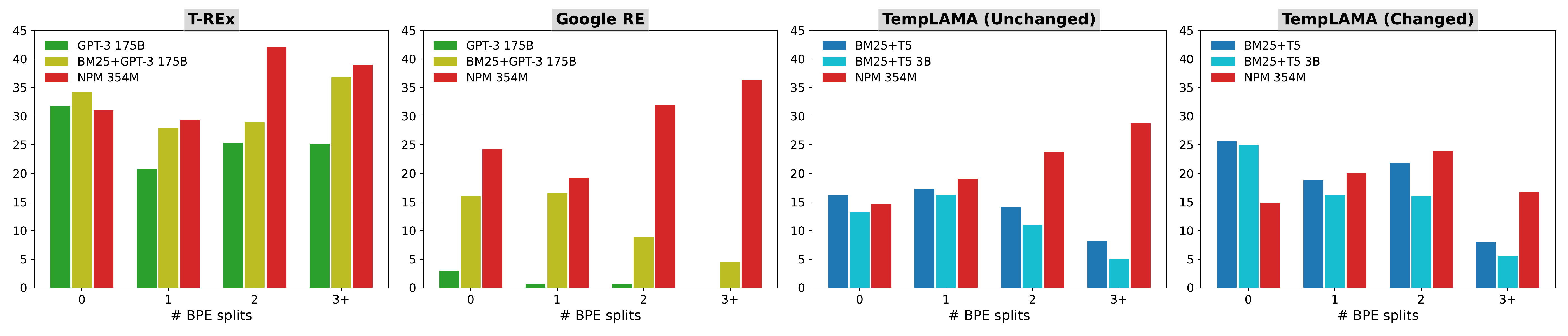}}\vspace{-.4em}
\caption{
    Performance on LAMA and TempLAMA tasks, broken down based on the number of BPE splits of the target entity, which is an indication of rarity of the entities (L:Frequent$\rightarrow$R:Rare).
    \ours\ outperforms GPT-3 or T5 more significantly when the target entities are rare.
}\label{fig:open-set-abl}
\end{figure*}

\begin{table}[!t]
    \centering \myfontsize
    \setlength{\tabcolsep}{4pt}
    \begin{tabular}{l @{\hspace{-1em}} r @{\hspace{1.5em}} cca}
        \toprule
            Model & \#Params & Unchanged & Changed & {AVG} \\
        \midrule
            \multicolumn{4}{l}{\textbf{\em Baselines}} \\
            T5          & 2.2x &               1.9 & 0.4 & 1.1 \\
            T5 3B       & 8.5x &               1.8 & 0.4 & 1.1 \\
            OPT 6.7B        & 19x   &          2.5 & 1.0 & 1.7 \\
            OPT 13B         & 37x   &          4.9 & 2.1 & 3.5 \\
        \cmidrule(lr){1-4}
            BM25 + T5 & 2.2x &      13.7$\rightarrow$14.9 & 3.0$\rightarrow$\textbf{20.1} & 17.5 \\
            BM25 + T5 3B & 8.5x &  11.9$\rightarrow$12.0 & 2.2$\rightarrow$17.8 & 14.9 \\
            BM25 + OPT 6.7B        & 19x & 10.2$\rightarrow$8.2 & 1.7$\rightarrow$11.3  & 9.7 \\
            BM25 + OPT 13B         & 37x & 14.8$\rightarrow$14.4 & 2.8$\rightarrow$16.6 & 15.5 \\
        \midrule
            \multicolumn{4}{l}{\textbf{\em Ours}} \\
            \ours  & 1.0x & 18.9$\rightarrow$\textbf{19.5} & 2.9$\rightarrow$17.5 & \textbf{18.5} \\
        \bottomrule
    \end{tabular}\vspace{-.1em}
    \caption{
        Results on \templama, on an unchanged set, a changed set, and a macro-average over two, respectively.
        xx$\rightarrow$xx indicates performance when using the outdated and the updated Wikipedia, respectively.
    }\label{tab:temporal-open-set}
\end{table}

\vspace{-.2em}
\paragraph{Results on temporal knowledge tasks.}

Table~\ref{tab:temporal-open-set} reports results on TempLAMA.
\ours\ retains its performance on the unchanged set ($18.9\rightarrow$19.5) and successfully updates its answers on the changed set ($2.9\rightarrow17.5$).
Its performance is significantly better than the performance of parametric models with up to 13B parameters, and is on par with a larger model with the retrieve-and-generate approach, which also successfully updates its answer by leveraging the updated corpus.
This is in agreement with prior work that shows the model with a \np\ component adapts to temporal updates by replacing the reference corpus at test time~\citep{izacard2022few}.
Nonetheless, the retrieve-and-generate approach is still significantly worse than \ours\ when the target entities are rare, 
which we show in the next paragraph.

\begin{table}[t]
    \centering \myfontsize
    \setlength{\tabcolsep}{2.2pt}
    \begin{tabular}{
        l @{\hspace{-0.5em}} r @{\hspace{2em}} r @{\hspace{0.8em}} r @{\hspace{1.2em}} r
    }
        \toprule
            Model & \#Params & \#L & w/o BM25 & w/ BM25 \\
        \midrule
            \multicolumn{4}{l}{\textbf{\em Baselines, English-only}} \\
            T5          & 2.2x      &&0.2 & 1.9\\
            T5 3B       & 8.5x      &&0.5 & 4.4\\
            OPT 6.7B    & 19x       &&0.4 & 22.3 \\
            OPT 13B     & 37x       &&1.0 & 24.6\\
        \midrule
            \multicolumn{4}{l}{\textbf{\em Ours, English-only}} \\
            \ours  & 1.0x           && \multicolumn{2}{c}{~~~~~~~~\textbf{52.4}} \\
        \midrule
            \multicolumn{4}{l}{\textbf{\em References, Multilingual}} \\
            mT5         & 3.4x & 101    &1.3 & 19.0\\
            mT5 XL      & 11x & 101     &4.1 & \best{56.6}\\
            BLOOM 3B & 8.5x & 46        &0.0 & 17.4\\
            BLOOM 7.1B & 20x & 46       &0.1 & 26.0 \\
        \bottomrule
    \end{tabular}\vspace{-.1em}
    \caption{
        Results on the entity translation task. See Table~\ref{tab:entity-translation} in Appendix~\ref{app:additional-results} for per-language results.
        {\em \#L} indicates the number of languages multilingual models are trained on.
        \textbf{Bold} and \best{Bold} indicate the best among monolingual models and the best including multilingual models, respectively.
        {\ours\ significantly outperforms all existing monolingual models, and approaches or outperforms larger multilingual models.}
    }\label{tab:entity-translation-summary}
\end{table}

\vspace{-.2em}
\paragraph{Performance on rare entities.}

We break down the instances on LAMA and TempLAMA based on the number of BPE splits of the target entity, e.g., {\em Thessaloniki} 
is one word that is split into 4 BPE tokens, thus the number of splits is 3. Since BPE splits a word if they are rare, the number of BPE splits indicates the rarity of the entity. We compare \ours\ with GPT-3 and BM25+GPT-3 on LAMA, and BM25+T5 (770M and 3B) on TempLAMA, the two most competitive baselines on each dataset.

Figure~\ref{fig:open-set-abl} reports results. On LAMA, \ours\ outperforms GPT-3 fairly consistently, with larger gains as the number of BPE splits increases.
On TempLAMA, while BM25+T5 is competitive on frequent entities with zero BPE split, it consistently lags behind \ours\ with $\geq1$ BPE splits.
This suggests that \ours\ is particularly good at addressing rare entities, compared to not only parametric models without retrieval but also the retrieve-and-generate approach.

\myskip{
\begin{table*}[t]
    \centering \myfontsize
    \setlength{\tabcolsep}{2.2pt}
    \begin{tabular}{l @{\hspace{-0.5em}} r @{\hspace{0.8em}} r @{\hspace{0.6em}} rrr rr rrr rrr rrr ra}
        \toprule
            Model & \#Params & \#L &
            \Ara & \Cze & \Gre & \Hin & \Heb & \Jap & \Kor & \Mal & \Mon & \Pol & \Rus & \Tal & \Tha & \Tur & \Chi & AVG \\
        \midrule
            \multicolumn{16}{l}{\textbf{\em Baselines, English-only}} \\
            T5          & 2.2x &&0.0&0.0&0.0&0.0&0.1&0.2&0.0&0.0&0.0&1.1&0.9&0.0&0.0&0.0&0.0&0.2 \\
            T5 3B       & 8.5x && 0.0 & 0.0 & 0.0&0.0&0.0&0.0&0.0&0.0&0.0&5.6&1.3&0.0&0.0&0.0& 0.0 & 0.5 \\
            OPT 6.7B    & 19x       &&0.0&0.0&0.3&0.0&0.0&0.0&3.1&0.0&0.0&0.0&2.9&0.0&0.0&0.0&0.0&0.4 \\
            OPT 13B     & 37x       &&1.5&0.0&1.2&0.7&0.0&0.0&1.4&0.0&0.0&1.1&7.4&0.0&0.0&1.3&0.1&1.0 \\
        \cmidrule(lr){1-19}
            BM25 + T5 & 2.2x &&0.0&5.5&0.3&0.2&0.5&0.0&0.2&1.9&0.0&6.8&0.8&1.2&0.0&11.3&0.0&1.9 \\
            BM25 + T5 3B & 8.5x && 0.0&12.8&0.1&0.7&0.2&0.8&0.0&0.0&1.6&28.8&1.7&0.0&0.0&20.0&0.0&4.4 \\
            BM25 + OPT 6.7B  & 19x   &&26.4&\textbf{54.1}&15.5&11.2&11.8&14.4&19.6&5.7&3.1&\textbf{47.5}&52.5&6.2&12.2&32.0&22.7&22.3 \\
            BM25 + OPT 13B  & 37x   &&17.3&51.4&24.9&15.5&27.8&12.3&22.0&11.3&7.8&45.8&48.2&8.8&18.9&\textbf{34.0}&23.3&24.6 \\
        \midrule
            \multicolumn{16}{l}{\textbf{\em Ours, English-only}} \\
            \ours  & 1.0x &&
            \textbf{51.9} & 33.0 & \best{60.9} & \textbf{63.2} & \best{63.7} & \best{59.0} & \textbf{60.5} & \best{50.9} & \textbf{46.9} & 33.3 & \best{61.2} & \best{51.2} & \best{60.8} & 32.7 & \textbf{56.9} & \textbf{52.4 }\\
        \midrule
            \multicolumn{16}{l}{\textbf{\em References, Multilingual}} \\
            mT5         & 3.4x & 101 &0.3&1.8&1.5&0.0&0.4&1.9&0.7&0.0&0.0&1.1&4.6&2.5&1.4&3.3&0.7&1.3 \\
            mT5 XL      & 11x & 101 &4.4&3.7&4.9&6.8&0.7&2.3&4.1&1.9&4.7&5.6&8.0&5.0&0.0&6.7&2.8&4.1 \\
            BLOOM 3B & 8.5x & 46 &0.0&0.0&0.0&0.0&0.0&0.0&0.0&0.0&0.0&0.0&0.0&0.0&0.0&0.0&0.3&0.0 \\
            BLOOM 7.1B & 20x & 46 &0.0&0.9&0.0&0.0&0.0&0.0&0.0&0.0&0.0&0.0&0.0&0.0&0.0&0.0&0.5&0.1 \\
        \cmidrule(lr){1-19}
            BM25 + mT5 & 3.4x & 101 &12.4&22.9&21.6&9.8&12.5&28.9&19.1&11.3&18.8&15.8&16.0&17.5&28.4&16.7&33.4&19.0 \\
            BM25 + mT5 XL & 11x & 101 &\best{64.4}&\best{64.2}&54.3&\best{65.6}&62.7&55.4&\best{69.4}&43.4&\best{62.5}&52.0&53.7&37.5&50.0&\best{48.7}&\best{65.0}&\best{56.6} \\
            BM25 + BLOOM 3B & 8.5x & 46 &24.2&25.7&1.7&13.3&15.1&18.5&17.9&5.7&6.2&21.5&11.1&10.0&27.0&18.0&44.5&17.4 \\
            BM25 + BLOOM 7.1B & 20x & 46 &19.0&49.5&11.4&20.8&8.1&30.1&25.4&5.7&6.2&\best{54.2}&29.0&6.2&37.8&33.3&53.7&26.0 \\
        \bottomrule
    \end{tabular}\vspace{-.1em}
    \caption{
        Results on the entity translation task.
        {\em \#L} indicates the number of languages multilingual models are trained on.
        \textbf{Bold} and \best{Bold} indicate the best among monolingual models and the best including multilingual models, respectively.
        See Table~\ref{tab:entity-translation-oracle} in Appendix~\ref{app:additional-results} for results with oracle passages.
        {\ours\ significantly outperforms all existing monolingual models, and approaches or outperforms larger multilingual models.}
    }\label{tab:entity-translation}
\end{table*}

\vspace{-.2em}
\paragraph{Results in Entity Translation.} 
Results on the entity translation task are shown in Table~\ref{tab:entity-translation}.
First, all baselines, including T5 and OPT, struggle to perform the task.
This is true even when the input is prepended with BM25 retrieval, or even with the oracle context (the paragraph that is guaranteed to contain the translation information; shown in Table~\ref{tab:entity-translation-oracle} of Appendix~\ref{app:additional-results}).
There are a couple of languages where models achieve moderate performance, e.g. Czech, Polish and Turkish. These are languages that are derived from Latin, thus the model primarily trained on the English data may be able to generalize to some extent.
However, for other languages, all models performs poorly.
This indicates that the models that directly estimate the probability of the vocabulary are not able to generate words that are extremely rare or unseen.
In contrast, \ours\ performs well across all languages, including more challenging cases such as Malayalam, Mongolian, Thai and Tamil.


In order to better calibrate performance of \ours, we provide reference performance of models that are purposely trained on the multilingual data.
We include mT5~\citep{xue2021mt5} and BLOOM~\citep{scao2022bloom}, trained on 101 languages and 46 languages, respectively.
\ours\ outperforms 3.4x larger mT5 and 20x larger BLOOM, and approaches 11x larger mT5, even though it is trained on English.

We think strong cross-lingual transferability of \ours\ is likely because it can retrieve a \phrase\ based on its surrounding context, even if it has not seen the exact word during training.
We think this is a unique property of our model, which uses a fully \np\ distribution instead of a softmax over the vocabulary.
%
This also suggests that an extension of \np\ models to a multilingual setting is promising. 

}

\vspace{-.2em}
\paragraph{Results in Entity Translation.}
Results on the entity translation task are shown in Table~\ref{tab:entity-translation-summary} (per-language results are reported in Table~\ref{tab:entity-translation} of Appendix~\ref{app:additional-results}).
T5 and OPT struggle to perform the task, both with and without BM25 retrieval.
In contrast, \ours\ performs well across all languages.

In order to better calibrate performance of \ours, we provide reference performance of models that are purposely trained on the multilingual data---mT5~\citep{xue2021mt5} and BLOOM~\citep{scao2022bloom}.
\ours\ outperforms 3.4x larger mT5 and 20x larger BLOOM, and approaches 11x larger mT5, even though it is trained on English.
We think strong cross-lingual transferability of \ours\ is likely because it can retrieve a \phrase\ based on its surrounding context, even if it has not seen the exact word during training.
\section{Conclusion}\label{sec:concl}We introduced \ours, a \np\ masked language model that replaces a softmax over the output vocabulary with a \np\ distribution over a reference corpus.
\ours\ can be efficiently trained using a contrastive objective and an in-batch approximation to a full corpus.
Zero-shot evaluation on 16 tasks shows that \ours\ outperforms significantly larger parametric models.
\ours\ is particularly good at rare patterns (word senses or facts), scaling and updating at test time, and predicting extremely rare if not unseen characters.

\section*{Limitation}\label{sec:limit}\paragraph{Scaling through the inference corpus.}
The size of the reference corpus is an additional dimension for model scale in \np\ models.
In this paper, we scale the corpus up to nearly 1B tokens, which is still smaller than the training data of very large language models~\citep{brown2020language,rae2021scaling}.
We think future work can scale it further using tools such as Distributed FAISS~\citep{johnson2019billion} or ScaNN~\citep{avq_2020}.

\paragraph{Significant memory usage.}
Using \ours\ saves GPU compute and memory compared to using models with more parameters.
However, \ours\ requires more RAM and disk memory due to embeddings of a reference corpus. For instance, the largest corpus in our experiments (full English Wikipedia) requires 70GB of RAM and 1.4TB of disk memory.
Future work can build more efficient \ours\ as done in prior work in nearest neighbor search~\citep{jegou2010product,norouzi2012fast,ge2014optimized,izacard2020memory,yamada-etal-2021-efficient}.

\vspace{-.2em}
\paragraph{Exploration of larger vocabulary.}
Large vocabulary is known to lead performance gains~\citep{conneau2020unsupervised} but is bounded in memory costs. Previous work explored more efficient softmax approximations~\citep{morin2005hierarchical,chen2015strategies,pmlr-v70-grave17a}.
Our \np\ training offers an alternative by removing the softmax over the vocabulary.
With the RoBERTa architecture, increasing the vocab size by 2x makes the baseline training 50\% more memory expensive, but does not increase the memory in training \ours.
However, this paper does not include more systematic evaluation on the effect of large vocabulary.
Future work can explore training \ours\ with a significantly larger vocabulary to further boost performance.

\vspace{-.2em}
\paragraph{Extension for generation.}
Our paper evaluates \ours\ only on prediction tasks.
It is currently non-trivial to use \ours\ for generation, since it is the encoder-only model. Future work can explore autoregressive generation as done in \citet{patel2022bidirectional} or use \ours\ for editing~\citep{schick2022peer,gao2022attributed}.

\vspace{-.2em}
\paragraph{Extension to few-shot learning and fine-tuning.}
Our paper focuses on zero-shot evaluation only.
Future work can extend \ours\ to a few-shot learning setup. In fact, fine-tuning \ours\ is significantly easier than fine-tuning larger models such as T5, OPT and GPT-3 which we compare \ours\ with, and can be explored in future work.

\vspace{-.2em}
\paragraph{Better cross-lingual transfer.}
Our work explored cross-lingual transfer in a limited setup where the model is trained on monolingual data.
We think future work can train multilingual \ours, and explore more comprehensive cross-lingual evaluation.
In fact, \np\ training may alleviate the burden of collecting large-scale multilingual corpora since it makes the model less sensitive to the language coverage in the training data, and may lead to significantly better cross-lingual transfer, as we demonstrate in the entity translation task.

\vspace{-.2em}
\paragraph{Limitation in speed.}
We find that search makes inference considerably slower than the counterpart without search.
We think that (1) search can significantly be faster with better engineering (we use the default hyperparameters of the FAISS index with no tuning) or better index, and
(2) the speed of \ours\ is still on par with the speed of significantly larger parametric models that \ours\ outperforms (see Table~\ref{tab:inference-speed}).
Moreover, while not explored in this work, there has been work that improves inference speed~\citep{he-etal-2021-efficient,alon2022neuro} that can be applied to \ours.
We leave improving inference speed to future work.

\begin{table}[t]
    \centering \myfontsize
    \setlength{\tabcolsep}{4pt}
    \begin{tabular}{l @{\hspace{0em}} r c c r r}
        \toprule
            Model & \#Params & FS & SP & Acc & \#Q/sec \\
        \midrule
            RoBERTa & 1.0x & & & 67.6 & 36.36 \\
            \ours$^\ddagger$ & 1.0x & \checkmark && 75.5 & 7.63 \\
        \midrule
            OPT 2.7B & 7.6x &&& 2.1 & 0.71 \\
            OPT 2.7B + BM25$^\ddagger$ & 7.6x && \checkmark & 8.3 & 0.28 \\
            OPT 6.7B & 19x &&& 4.2 & 0.18 \\
            OPT 6.7B + BM25$^\ddagger$ & 19x && \checkmark & 10.7 & 0.12 \\
            \ours$^\ddagger$ & 1.0x && \checkmark & 10.8 & 4.52 \\
        \bottomrule
    \end{tabular}\vspace{-.1em}
    \caption{
        Inference speed measured on Subj with $|\mathcal{C}|=15$M (the first block) and NQ with $|\mathcal{C}|=810$M (the second block).
        A single GPU used (Quadro GP100).
        $\ddagger$ indicates the corpus is used.
        `FS' and `SP' indicate that a FAISS index is used and a sparse index (+ exact inner product search in case of \ours) is used, respectively.
        \ours\ is slower than the same-sized parametric model, but is faster than larger models (either with or without retrieval) while outperforming or matching performance.
    }\label{tab:inference-speed}
\end{table}

\section*{Acknowledgements}
We thank Ari Holtzman, Eric Wallace, Iz Beltagy, Jinhyuk Lee, Jungsoo Park, Mark Johnson, Noah Smith, Ofir Press, Patrick Lewis, Xiang Deng, Xinxi Lyu, Zexuan Zhong, UW-NLP members and anonymous reviewers for discussion and comments on the paper.
This research was supported by NSF IIS-2044660, ONR N00014-18-1-2826, ONR MURI N00014- 18-1-2670, an Allen Distinguished Award and gifts from AI2. SM is supported by a J.P. Morgan fellowship.

\bibliography{abbr,acl}
\bibliographystyle{acl_natbib}

\clearpage
\appendix
\section{Model Details}\label{app:model-details}\myskip{
\subsection{Training objective of \oursmulti}\label{app:ours-multi-objective}

Consider the $i$-th sequence in the batch.
Let $\hat{\mathcal{X}} = \{(\hat{x}^i_{t_j}, \hat{x}^i_{t_j+1})\}_{j=1}^M$ be a list of $M$ masked spans (represented as two consecutive \mask\ tokens),
and $\gx = x_{g'(t_j)}...x_{g'(t_j)+l^i_j-1}$ be the the corresponding \ngram\ of the $j$-th masked span, i.e., original tokens that were replaced by two \mask\ tokens $\hat{x}^i_{t_j}$ and $\hat{x}^i_{t_j+1}$, whose length is $l^i_j$.
The training objective for this $i$-th sequence is defined as$$
    \sum_{j=1}^M l_\mathrm{s}(\hat{x}^i_{t_j},\gx) + l_\mathrm{e}(\hat{x}^i_{t_j+1},\gx).
$$
$l_\mathrm{s}(\hat{x}^i_{t_j},\gx)$ and 
$l_\mathrm{e}(\hat{x}^i_{t_j+1},\gx)$ are respectively defined as
\begingroup\makeatletter\def\f@size{10}\check@mathfonts
\begin{eqnarray*}
    -\mathrm{log}\frac{
        \sum_{\mathbf{y} \in \ysPos} \mathrm{exp}(\mathrm{sim}(\mathbf{\hat{x}}_{i_j}, \mathbf{y}))
    }{
        \sum_{\mathbf{y} \in \ysPos \cup \ysNeg} \mathrm{exp}(\mathrm{sim}(\mathbf{\hat{x}}_{i_j}, \mathbf{y}))
    }, \\
    -\mathrm{log}\frac{
        \sum_{\mathbf{y} \in \yePos} \mathrm{exp}(\mathrm{sim}(\mathbf{\hat{x}}_{i_j+1}, \mathbf{y}))
    }{
        \sum_{\mathbf{y} \in \yePos \cup \yeNeg} \mathrm{exp}(\mathrm{sim}(\mathbf{\hat{x}}_{i_j+1}, \mathbf{y}))
    }.
\end{eqnarray*} \endgroup

\vspace{-.15em}
\paragraph{In-batch positives and negatives.}
Positives and negatives are tokens from other sequences in the batch that matches the corresponding \ngram.
For instance, an $m$-th token in the $j$-th sequence, $x^j_m$, is included in $\ysPos$ if $x^j_m...x^j_{m+l^i_j-1}$ is identical to $\gx$ and $i\neq j$.
Putting everything together, $\ysPos$, $\ysNeg$, $\yePos$ and $\yeNeg$ are respectively defined as:
\begingroup\makeatletter\def\f@size{10}\check@mathfonts
\begin{eqnarray*}
    \big\{ x^j_m | \gx=x^j_m...x^j_{m+l^i_j-1} \text{~and~} i \neq j\big\}, \\
    \big\{ x^j_m | \gx \neq x^j_m...x^j_{m+l^i_j-1} \text{~and~} i \neq j\big\}, \\
    \big\{ x^j_m | \gx=x^j_{m-l^i_j+1}...x^j_m \text{~and~} i \neq j\big\}, \\
    \big\{ x^j_m | \gx \neq x^j_{m-l^i_j+1}...x^j_m \text{~and~} i \neq j\big\}.
\end{eqnarray*} \endgroup

\paragraph{Approximation in \oursmulti\ inference.}
As described in Section~\ref{subsubsec:ours-inference-multi}, we predict the \ngram\ by (1) assigning a score $s(c_{i_1}...c_{i_n})$ to each distinct \ngram\ $c_{i_1}...c_{i_n}$ in the corpus:
$$\mathrm{exp}(\mathrm{sim}(\mathbf{q}^\mathrm{start},\mathbf{c}_{i_1}))+\mathrm{exp}(\mathrm{sim}(\mathbf{q}^\mathrm{end}, \mathbf{c}_{i_n})),$$
and taking an argmax of the aggregation over them:
$$\argmax_{v^* \in \mathcal{V}^*} \sum_{c^* \in \mathcal{C}^*}\mathbb{I}[v^*=c^*]s(c^*),$$ where $\mathcal{V}^*$ and $\mathcal{C}^*$ are sets of any possible \ngrams\ defined by the vocabulary $\mathcal{V}$ and the corpus $\mathcal{C}$, respectively.
In practice, since computing scores over all possible \ngrams\ in $\mathcal{C}$ is infeasible, an approximation is made by finding the top $k$ {\em start} tokens and the top $k$ {\em end} tokens, and computing scores over \ngrams\ composed by these tokens only. More precisely,\begin{eqnarray*}
    s^1,s^2,\cdots,s^k &=& \argtopk_s\mathrm{sim}(\mathbf{q}^\mathrm{start},\mathbf{c}_{s}), \\
    e^1,e^2,\cdots,e^k &=& \argtopk_e\mathrm{sim}(\mathbf{q}^\mathrm{end},\mathbf{c}_{e}) \\
\end{eqnarray*} are obtained by using a FAISS index. We then define a set of candidate \ngrams\ $\mathcal{C}^*_\mathrm{cand}$ as:
\begingroup\makeatletter\def\f@size{10}\check@mathfonts
$$
    \left(
        \bigcup\limits_{i=1}^k \bigcup\limits_{j=1}^{l_\mathrm{max}} c_{s_i}...c_{s_i+j-1}
    \right)
    \cup
    \left(
        \bigcup\limits_{i=1}^k \bigcup\limits_{j=1}^{l_\mathrm{max}} c_{e_i-j+1}...c_{e_i}
    \right),
$$ \endgroup
and then return the following as the prediction:$$\argmax_{v^* \in \mathcal{V}^*} \sum_{c^* \in \mathcal{C}^*_\mathrm{cand}}\mathbb{I}[v^*=c^*]s(c^*).$$
Note that we do not need to build the index for the start and the end separately. We use one index as in \ours, because it indexes vectors from the corpus that are independent from the start and the end.

}

\subsection{Details of \ours}\label{app:model-details-ours}
\paragraph{Approximation at inference.}
Given $\mathbf{q}^\mathrm{start}$ and $\mathbf{q}^\mathrm{end}$,
we take the top $k$ tokens with the highest similarity scores with each of them, and compute scores over spans composed by these tokens.
Let $c_{i:j}^*$ be a span in $\mathcal{C}$ from the $i$-th token to the $j$-th token, and $\mathrm{E}(c) \in \mathbf{R}^h$ be a vector corresponding to a token $c \in \mathcal{C}$.
We find the top $k$ tokens for the start and the end:\begin{eqnarray*}
    c_{\mathrm{s}_1},c_{\mathrm{s}_2},\cdots,c_{\mathrm{s}_k} &=& \argtopk_{c \in \mathcal{C}}\mathrm{sim}(\mathbf{q}^\mathrm{start},\mathrm{E(c)}), \\
    c_{\mathrm{e}_1},c_{\mathrm{e}_2},\cdots,c_{\mathrm{e}_k} &=& \argtopk_{c \in \mathcal{C}}\mathrm{sim}(\mathbf{q}^\mathrm{end},\mathrm{E(c)})
\end{eqnarray*} using a fast nearest neighbor search.
We then define a set of candidate \phrases\ $\tilde{\mathcal{C}}^*$ as:
\begingroup\makeatletter\def\f@size{10}\check@mathfonts
$$
    \left(
        \bigcup\limits_{i=1}^k \bigcup\limits_{j=1}^{l_\mathrm{max}} c_{\mathrm{s}_i:\mathrm{s}_i+j-1}^*
    \right)
    \cup
    \left(
        \bigcup\limits_{i=1}^k \bigcup\limits_{j=1}^{l_\mathrm{max}} c_{\mathrm{e}_i-j+1:\mathrm{e}_i}^*
    \right),
$$ \endgroup
and predict:$$\argmax_{v^* \in \mathcal{V}^*} \sum_{c^* \in \tilde{\mathcal{C}}^*}\mathbb{I}[v^*=c^*]\mathrm{exp}\mathrm{sim}(\mathbf{q}, \mathrm{E}(c^*)),$$where $\mathrm{E}(c^*) \in \mathbf{R}^{2h}$ is a vector corresponding to $c^*$, and $\mathcal{V}^*$ is a set of any possible \ngrams\ defined by the vocabulary $\mathcal{V}$.

\subsection{Training Details}\label{app:impl-details}

All implementation was done with PyTorch~\citep{paszke2019pytorch}, PyTorch Lightning\footnote{\url{https://github.com/Lightning-AI/lightning}} and Huggingface Transformers~\citep{wolf-etal-2020-transformers}.

\paragraph{Masking.}
We use a masking ratio of 15\% for all models, following the standard in prior work~\citep{devlin2019bert,liu2019roberta,joshi2020spanbert}.
%
%
We implement masking as follows: (1) we first identify all possible candidate spans (spans that positives are found from other sequences in the batch), (2) sample the length of spans to mask from a geometric distribution with a hyperparameter $p=0.5$, and (3) mask the spans with respect to the sampled length until the masking budget has been spent.
We do not mask more than 128 spans from one sequence, and do not mask the span if the same span has been masked for more than ten times within the batch in order to prevent repeatedly masking overly frequent spans.

For \maskS\ and \maskE, we use the \mask\ vocab from the RoBERTa tokenizer. Note that it is not necessary to use different tokens for \maskS\ and \maskE\ since the Transformer can handle positional information.

\subsection{A special case: \ourssingle}\label{app:model-details-ours-multi}
Along with \ours, we introduce \textbf{\ourssingle}, which outputs a \np\ distribution over every single {\em token} in $\mathcal{C}$, instead of a {\em \phrase}.
To some extent, \ours\ is a strict generalization of \ourssingle, and \ourssingle\ still has a problem that existing encoder-only models have, e.g., can only fill in the \mask\ with a single token.
We however think \ourssingle\
can be useful for some applications, e.g., for fine-tuning, as existing encoder-only models are used for. 

\paragraph{Inference.}

Given a reference corpus $\mathcal{C} = \{c_1, \cdots, c_N \}$,
we construct $N$ number of $h$-dimensional vectors $\mathbf{c}_1, \cdots, \mathbf{c}_N \in \mathbb{R}^h$ by feeding the text into the encoder.
At inference time, given a query whose $t$-th token is \mask, we feed it into the encoder:$$\mathbf{q}_1...\mathbf{q}_L = \mathrm{Encoder}(q_1...q_{t-1}, \mask, q_{t+1}...q_L).$$
We take $\mathbf{q}_t$ as a vector that represents the \mask\ token in the query.
Finally, the prediction is made by aggregating the similarity scores to the tokens in $\mathcal{C}$:$$
    \argmax_{v \in \mathcal{V}} \sum_{c \in \mathcal{C}} \mathbb{I}[c=v] \mathrm{exp}(\mathrm{sim}(\mathbf{q}_t, \mathrm{E}(c))),
$$where $\mathrm{E}(c) \in \mathbf{R}^h$ is a vector corresponding to $c$, and $\mathcal{V}$ is the vocabulary set.

In practice, since computing scores over all tokens in $\mathcal{C}$ is infeasible, an approximation is made by computing scores for the top $k$ nearest neighbors only, and treating other tokens to have a similarity score of $-\mathrm{Inf}$. More precisely:$$c^1,c^2,\cdots,c^k=\argtopk_{c \in \mathcal{C}}\mathrm{sim}(\mathbf{q}_t, \mathrm{E}(c))$$ are obtained by using an index (e.g., FAISS~\citep{johnson2019billion}), and the following is returned as a prediction:$$\argmax_{v \in \mathcal{V}} \sum_{1 \leq i \leq k} \mathbb{I}[c^i=v] \mathrm{exp}(\mathrm{sim}(\mathbf{q}_t, \mathrm{E}(c^i))).
$$

\vspace{-.2em}
\paragraph{Training.}

Let $x_1^i...x_L^i$ be the $i$-th sequence in the batch, whose subset is replaced with \mask\ and converted to $\hat{x}_1^i...\hat{x}_L^i$. 
Both the unmasked sequence and the masked sequence are fed into the encoder, and each token is mapped into an $h$-dimensional vector:
\begin{eqnarray*}
    \mathbf{x}_1^i \cdots \mathbf{x}_L^i&=&\mathrm{Encoder}(x_1^i \cdots x_L^i), \\
    \mathbf{\hat{x}}_1^i \cdots \mathbf{\hat{x}}_L^i&=&\mathrm{Encoder}(\hat{x}_1^i \cdots \hat{x}_L^i).
\end{eqnarray*}
The training objective is then defined as:$$
    \sum_{t=1}^L \mathbb{I}[\hat{x}_t=\mask]l(x_t^i,\hat{x}_t^i),
$$ where $l(x_t^i,\hat{x}_t^i)$ is $$-\mathrm{log}\frac{
        \sum_{\mathbf{y} \in \mathcal{Y}^+(x_t^i)} \mathrm{exp}(\mathrm{sim}(\mathbf{\hat{x}}_t^i, \mathbf{y}))
    }{
        \sum_{\mathbf{y} \in \mathcal{Y}^+(x_t^i) \cup \mathcal{Y}^-(x_t^i)} \mathrm{exp}(\mathrm{sim}(\mathbf{\hat{x}}_t^i, \mathbf{y}))
    }.
$$

Here,
$\mathrm{sim}(\cdot,\cdot)$ is a similarity function defined in Section~\ref{subsec:ours-inference},
and $\mathcal{Y}^+(x_t^i)$ and $\mathcal{Y}^-(x_t^i)$ are {\em positives} and {\em negatives} of $x_t^i$---tokens from {\em other} sequences in the batch that share and do not the vocab, respectively.\begin{eqnarray*}
    \mathcal{Y}^+(x^i_t)&=&\left\{ x^j_m | x^i_t = x^j_m \text{~and~} i \neq j\right\}, \\
    \mathcal{Y}^-(x^i_t)&=&\left\{ x^j_m | x^i_t \neq x^j_m \text{~and~} i \neq j\right\}.
\end{eqnarray*}

\begin{table*}[t]
    \centering \myfontsize
    \setlength{\tabcolsep}{4pt}
    \begin{tabular}{l @{\hspace{0em}} r @{\hspace{0.8em}} r @{\hspace{0.5em}} rl}
        \toprule
            Dataset & $|\mathcal{D}|$ &  $|\mathcal{D}_\mathrm{s}|$ & \# labels & Example \\
        \midrule
            \multicolumn{5}{l}{\textbf{\em Closed-set tasks}} \\
            AGN  & 120,000 & 3,000 & 4  &
            \makecell[l]{Indiana defends its NCAA mens's soccer title by edging UC Santa Barbara in penalty kicks. \\ The text topic is about \mask. (\mask=\{politics, sports, business, technology\})}
            \\
        \cmidrule(lr){5-5}
            Yahoo   & 60,000 & 3,000 & 10  & \makecell[l]{Company for cemecal at espaniea? Answer: Can you give us more info? The text topic is \\ about \mask. (\mask=\{society, science, health,
            education, computer, sports, business, \\ entertainment, family, politics\})} \\
        \cmidrule(lr){5-5}
            Subj    & 2,000 & 2,000 & 2 & \makecell[l]{He tells mitchell that he is now in debt. This is a \mask. \\
            (\mask=\{review, summary\})} \\
        \cmidrule(lr){5-5}
            SST-2   & 2,210 & 2,210 & 2 & It was \mask. (\mask=\{great, terrible\})\\
        \cmidrule(lr){5-5}
            MR      & 2,000 & 2,000 & 2 & Simplistic, silly and tedious. It was \mask.
            (\mask=\{great, terrible\})\\
        \cmidrule(lr){5-5}
            RT      & 1,066 & 1,066 & 2 & weird. rewarding. It was \mask. (\mask=\{great, terrible\})\\
        \cmidrule(lr){5-5}
            CR      & 2,000 & 2,000 & 2 & I am very pleased so far. It was \mask. (\mask=\{great, terrible\})\\
        \cmidrule(lr){5-5}
            Amz     & 400,000 & 3,000 & 2 & It was \mask. (\mask=\{great, terrible\})\\
        \cmidrule(lr){5-5}
            RTE     & 277 & 277 & 2 & \makecell[l]{Most commercial logwood is grown in Honduras, right? \mask, plants are grown in water \\ or in substance other than soil. (\mask=\{Yes, No\})} \\
        \midrule
            \multicolumn{5}{l}{\textbf{\em Open-set tasks}} \\
            LAMA T-REx       & 34,039 & 2,983 & -   & AVCDH is owned by \mask. \\
            LAMA Google RE   & 5,200 & 1,856 & -    & Joshua Mathiot died in \mask. \\
            KAMEL            & 46,800 & 3,000 & -   & What is followed by So-Lo? Answer: \mask. \\
            NQ               & 3,610 & 3,000 & -    & who sang i ran all the way home? The answer is: \mask. \\
            TQA              & 11,313 & 3,000 & -   & Who wrote the opera Carmen? The answer is: \mask. \\
            \templama \\
            ~~~~- changed      & 3,360 & 3,000 & - & Contributor Covenant is developed by \mask. \\
            ~~~~- unchanged    & 3,360 & 3,000 & - & Atari 8-bit family is developed by \mask. \\
            Entity translation & 10,452 & 6,622 & -   & The Korean translation of Banpo Bridge is: \mask. \\
        \bottomrule
    \end{tabular}\vspace{-.1em}
    \caption{
        Statistics of downstream datasets.
        $|\mathcal{D}|$ and $|\mathcal{D}_\mathrm{s}|$ indicate the number of test examples on the original data and the subsampled data, respectively.
        See Appendix~\ref{app:eval-details} for details.
    }\label{tab:dataset-statistics}
\end{table*}

\begin{table*}[t]
    \centering \myfontsize
    \begin{tabular}{l l @{\hspace{1em}} r @{\hspace{1em}} l}
        \toprule
            Corpus name & Source & $|\mathcal{C}|$ & Datasets used \\
        \midrule
            En-Wiki+CCNews & Subset of En-Wiki 08/01/2019 and CCNews & 126M & AGN, Yahoo, RTE \\
            Subjectivity corpus & Raw IMDB & 15M & Subj \\
            Review corpus & Amazon and IMDB & 62M & SST-2, MR, RT, CR, Amz \\
            En-Wiki 2019 & En-Wiki 08/01/2019 & 810M & All open-set tasks \\
            En-Wiki 2022 & En-Wiki 08/01/2022 & 858M & \templama \\
        \bottomrule
    \end{tabular}\vspace{-.1em}
    \caption{
        Statistics of the retrieval corpus. $|\mathcal{C}|$ indicates the number of tokens in the corpus.
    }\label{tab:corpus-statistics}
\end{table*}

\subsection{Inference on closed-set tasks}\label{app:inference-closed-set}
When applying \ours\ and \ourssingle\ on closed-setk tasks, we closely follow \citet{shi2022nearest} who adapts \knnlm\ for zero-shot inference on classification tasks. We assume a fuzzy verbalizer: $f: \mathcal{Y}\rightarrow\tilde{\mathcal{V}}$, where  $\mathcal{Y}$ is a set of labels in the task and $\tilde{\mathcal{V}} \in \mathcal{V}$ is a subset of the vocabulary $\mathcal{V}$. The fuzzy verbalizer maps a label to a set of tokens that express the label, e.g., in a sentiment classification task, $f(\texttt{Positive})$ includes {\em awesome} or {\em great}, and $f(\texttt{Negative})$ includes {\em terrible} or {\em broken}.

\vspace{.3em}
\textbf{\ourssingle} is given a query vector $\mathbf{q} \in \mathbb{R}^h$ and predicts:$$
    \argmax_{y \in \mathcal{Y}} \sum_{c \in \mathcal{C}} \mathbb{I}[c \in f(y)] \mathrm{exp}\left(
        \frac{\mathrm{sim}(\mathbf{q}, \mathrm{E}(c))}{\tau}
    \right),
$$where $\mathrm{E}(c) \in \mathbf{R}^h$ is a vector corresponding to $c$, and $\tau$ is a hyperparameter.

\vspace{.3em}
\textbf{\ours} is given a query vector $\mathbf{q} \in \mathbb{R}^{2h}$ and predicts:$$
    \argmax_{y \in \mathcal{Y}} \sum_{c^* \in \mathcal{C}^*} \mathbb{I}[c^* \in f(y)] \mathrm{exp}\left(
        \frac{\mathrm{sim}(\mathbf{q}, \mathrm{E}(c^*))}{\tau}
    \right),
$$where $\mathrm{E}(c^*) \in \mathbf{R}^{2h}$ is a vector corresponding to $c^*$.
Note that this is essentially equivalent to\begin{eqnarray*}
    \argmax_{y \in \mathcal{Y}} \sum_{c \in \mathcal{C}} \mathbb{I}[c \in f(y)] \mathrm{exp}\Bigg(~~~~~~~~~~~~~~~~~~~~~~~~~~~~~~\\
    ~~~~~~~~~~\frac{\mathrm{sim}(\mathbf{q}^\mathrm{start}, \mathrm{E}(c))}{\tau}
    + \frac{\mathrm{sim}(\mathbf{q}^\mathrm{end}, \mathrm{E}(c))}{\tau}
    \Bigg).
\end{eqnarray*}
We use $\tau=5.0$ for both \ourssingle\ and \ours.

\section{Evaluation Details}\label{app:eval-details}Table~\ref{tab:dataset-statistics} reports statistics and templates on each downstream task, and Table~\ref{tab:corpus-statistics} reports statistics of the retrieval corpus used in experiments.

For closed-set tasks, we use templates and verbalizers provided by \citet{shi2022nearest} for most datasets, except two datasets.
For RTE, we use the template from \citet{artetxe2022role}.
For Subj, we write our own template, motivated by \citet{zhong2022describing} that found Subj is mainly about differentiating a review and a summary.
For open-set tasks, we use templates provided by the original authors, except NQ and TQA for which we use the templates from GPT-3~\citep{brown2020language}.
Due to limited computation resource, we subsample the data to include up to 3,000 examples, following the standard from prior work~\citep{zhao2021calibrate,shi2022nearest}.
For closed-set tasks, we use exactly the same set of data as \citet{shi2022nearest}, and for open-set tasks, we use the same script to subsample the data.
For LAMA T-REx and Google RE, we subsample up to 1,000 examples for each of $1$, $2$, $3$ and $4+$ grams.
For the entity translation task, we subsample up to 1,000 examples per language.

\vspace{.5em}
The following is a more detailed description of open-set tasks used in Section~\ref{sec:exp-open}.

\vspace{.5em}
\textbf{LAMA~\citep{petroni-etal-2019-language}}
is a factual probing benchmark that is designed to quantify the amount of factual knowledge in the model. It requires the model to predict the object given a subject-relation tuple in a cloze format.
We use two versions of LAMA~\citep{petroni-etal-2019-language}: (1) LAMA T-REx, derived from \citet{elsahar2018t} and (2) LAMA Google-RE, derived from the Google-RE corpus.\footnote{\url{https://code.google.com/archive/p/relation-extraction-corpus}}
For each version, we additionally consider the UHN (UnHelpfulNames) subset~\citep{Poerner2019BERTIN}) where instances whose subject strongly hints the object by names (e.g., \texttt{Apple Watch} and \texttt{Apple}) are excluded.
We also consider the hard subset of T-Rex from \citet{zhong2021factual}.

Note that \citet{petroni-etal-2019-language} only include triples whose object is one token based on BERT~\citep{devlin2019bert}; however, with a different pretrained model like RoBERTa, entities could be multiple BPE tokens. Entities that are splitted into multiple BPE tokens are more rare entities.

\vspace{.5em} \noindent
\textbf{KAMEL~\citep{kalo2022kamel}} is another factual probing task as LAMA but with a few key differences to make it more general and broad: (1) it includes a broader coverage of triples, (2) it removes the constraint that the object is one token based on BERT, (3) it includes objects with literal values, and (4) it has a question answering format. 

\vspace{.5em} \noindent
\textbf{Natural Questions} (NQ, \citet{kwiatkowski2019natural}) and \textbf{TriviaQA} (TQA, ~\citet{joshi-etal-2017-triviaqa}) are two welll-studied open-domain question answering datasets.
We use the open-version of NQ~\citep{lee2019latent} and TQA where the question is the only input and the model should use its knowledge to answer the question.

\vspace{.5em} \noindent
\templamabold is a task that requires probing knowledge with temporal updates.
The task is first introduced by \citet{dhingra2022time} and \citet{jang2022temporalwiki}; however, we could not use either of existing data as their time split do not match our training.
We therefore create the data by using a script provided by \citet{dhingra2022time} but using the 2019 and the 2022 dumps.
We take Wikipedia triples whose relations are available for a template from either \citet{petroni-etal-2019-language} or \citet{dhingra2022time}.
We then include triples whose object entities differ between the 2019 dump and the 2022 dump (due to the entity being updated), or only appear in the 2022 dump (due to the subject or the relation being added) to the {\em changed} set.
Otherwise, triples are included in the {\em unchanged} set. 
We additionally find that many triples are overly difficult because the fact is extremely niche and not really known. 
We thus filter the data to only include facts that appear in Wikipedia.
Specifically, we include triples if the subject has a corresponding Wikipedia page and the object entity appears in that Wikipedia page.

\begin{table}[t]
    \centering \small 
    \begin{tabular}{ll @{\hspace{2em}} rr}
        \toprule
            ISO Code & Language & $|\mathcal{D}|$ & $|\mathcal{D}_\mathrm{s}|$ \\
        \midrule
            \Chi & Chinese     & 3,199 & 1,000 \\
            \Ara & Arabic      & 2,013 & 1,000 \\
            \Gre & Greek       & 1,618 & 1,000 \\
            \Heb & Hebrew      & 841 & 841 \\
            \Rus & Russian     & 758 & 758 \\
            \Jap & Japanese    & 471 & 471 \\
            \Hin & Hindi       & 427 & 427 \\
            \Kor & Korean      & 418 & 418 \\
            \Pol & Polish      & 177 & 177 \\
            \Tur & Turkish     & 150 & 150 \\
            \Cze & Czech       & 109 & 109 \\
            \Tal & Tamil       & 80 & 80\\
            \Tha & Thai        & 74 & 74 \\
            \Mon & Mongolian   & 64 & 64 \\
            \Mal & Malayalam   & 53 & 53\\
        \midrule
            TOTAL && 10,452 & 6,622 \\
        \bottomrule
    \end{tabular}\vspace{-.1em}
    \caption{
        Statistics of the entity translation benchmark. Languages are sorted based on their availabilities.
    }\label{tab:entity-translation-statistics}
\end{table}

\begin{table*}[t]
    \centering \myfontsize
    \setlength{\tabcolsep}{5pt}
    \begin{tabular}{l @{\hspace{-0.2em}} rc @{\hspace{2em}}  rrrrrrrr}
        \toprule
            \multirow{2}{*}{Model} & \multirow{2}{*}{\#Params} & \multirow{2}{*}{$\mathcal{C}$} & \multicolumn{3}{c}{T-REx} & 
            \multicolumn{2}{c}{Google RE} &\multirow{2}{*}{KML} & \multirow{2}{*}{TQA} & \multirow{2}{*}{NQ} \\
            \cmidrule(lr){4-6} \cmidrule(lr){7-8}
            &&& All & UHN & Hard & All & UHN \\
        \midrule
            \multicolumn{6}{l}{\textbf{\em Baselines (encoder-decoder)}} \\
            T5          & 2.2x && 13.3 & 5.5 & 10.7 & 1.1 & 0.4 & 1.6 & 4.2 & 0.5 \\
            T5 3B       & 8.5x && 12.1 & 8.2 & 11.5 & 2.1 & 0.7 & 3.6 & 9.0 & 2.0 \\
        \cmidrule(lr){1-11}
            BM25 + T5 & 2.2x & \checkmark & 22.2 & 20.3 & 22.4 & 16.4 & 16.6 & 13.9 & 31.4 & 5.2 \\
            BM25 + T5 3B & 8.5x & \checkmark & 21.6 & 19.0 & 21.8 & 18.5 & 15.5 & \textbf{16.2} & 39.6 & 10.8 \\
        \midrule
            \multicolumn{6}{l}{\textbf{\em Baselines (decoder-only)}} \\
            OPT 2.7B  & 7.6x & & 9.8 & 6.7 & 8.3 & 0.0 & 0.0 & 1.6 & 9.9 & 2.1 \\
            GPT-3 2.7B  & 7.6x  && 4.4 & 2.6 & 3.8 & 0.0 & 0.0 & 2.1 & 5.2 & 1.1 \\
            OPT 6.7B  & 19x& & 11.6 & 9.9 & 10.7 & 0.6 & 0.3 & 3.2 & 20.9 & 4.2 \\
            GPT-3 6.7B  & 19x   && 8.1 & 5.0 & 6.7 & 0.0 & 0.0 & 2.1 & 12.4 & 3.1 \\
            OPT 13B  & 37x & & 15.0 & 12.7 & 12.7 & 0.3 & 0.3 & 2.5  & 22.5 & 4.2 \\
            GPT-3 13B   & 37x   && 16.4 & 13.7 & 15.5 & 0.8 & 0.4 & 2.2 & 25.5 & 5.2 \\
            GPT-3 175B  & 500x  && 25.7 & 24.1 & 24.7 & 1.1 & 1.0 & 6.5  & \textbf{49.0} & \textbf{11.4} \\
        \cmidrule(lr){1-11}
            BM25 + OPT 2.7B  & 7.6x& \checkmark & 14.8 & 14.1 & 13.8 & 4.4 & 3.7 & 11.3 & 28.5 & 8.3 \\
            BM25 + GPT-3 2.7B & 7.6x  & \checkmark & 3.5 & 3.4 & 3.6 & 0.1 & 0.1 & 5.2 & 14.5 & 6.1 \\
            BM25 + OPT 6.7B  & 19x& \checkmark & 14.8 & 14.3 & 14.9 & 4.1 & 3.3 & 8.2 & 29.9 & 10.7 \\
            BM25 + GPT-3 6.7B & 19x   & \checkmark & 14.9 & 15.3 & 15.1 & 4.4 & 3.5 & 7.0 & 21.1 & 8.8 \\
            BM25 + OPT 13B  & 37x & \checkmark & 18.9 & 19.1 & 19.3 & 3.8 & 3.1 & 10.6 & 34.0 & 10.7 \\
            BM25 + GPT-3 13B & 37x   & \checkmark & 22.2 & 22.7 & 22.4 & 11.8 & 11.2 & 8.9 & 32.4 & 11.2 \\
            BM25 + GPT-3 175B & 500x  & \checkmark & 32.0 & \textbf{31.6} & 31.3 & 11.4 & 11.9 & 12.2  & 44.9 & 6.4 \\
        \midrule
            \multicolumn{6}{l}{\textbf{\em Ours (encoder-only, nonparametric)}} \\
            \ours & 1.0x & \checkmark & \textbf{34.5} & 29.0 & \textbf{32.1} & \textbf{27.9} & \textbf{23.0} & 15.6 & 32.2 & 10.8 \\
        \bottomrule
    \end{tabular}\vspace{-.1em}
    \caption{
        Results on open-set tasks (numbers used in Figure~\ref{fig:open-set-results}).
        {\em \# Params} indicates the relative number of model parameters compared to RoBERTa large (354M), and $\mathcal{C}$ indicates whether a text corpus is used.
        For LAMA (T-REx and Google RE), the macro-averaged EM over 1, 2, 3 and 4+ grams are reported.
        All models are zero-shot.
        \ours\ significantly outperforms larger parameters models, either with and without a retrieval-and-generate approach that uses BM25.
    }\label{tab:open-set}
\end{table*}

\vspace{.5em} \noindent
\textbf{Entity translation} requires translating an entity from English to other languages that are not Latin based.
While this is mainly to evaluate if the model can generate rare or unseen characters that are not in English, the entity translation task itself is a vital and challenging task in real applications such as machine translation~\citep{Babych2003ImprovingMT,yan2018impact} and cross-lingual question answering~\citep{tydiqa,xorqa}.
It is often beyond a series of simple translations of each word, or spelling out its pronunciation~\citep{moore2003learning,hassan2007improving,Sun2017CrossLingualEA}. For instance, the Korean translation of {\em Banpo Bridge} in Figure~\ref{fig:intro} (\begin{CJK}{UTF8}{mj}반포대교\end{CJK}) is not the concatenation of the translations of {\em Banpo} and {\em Bridge} (\begin{CJK}{UTF8}{mj}반포 다리\end{CJK}). 

We first identify a list of 15 non-Latin languages: Arabic (\Ara), Czech (\Cze), Greek (\Gre), Hindi (\Hin), Hebrew (\Heb), Japanese (\Jap), Korean (\Kor), Malayalam (\Mal), Mongolian (\Mon), Polish (\Pol), Russian (\Rus), Tamil (\Tal), Thai (\Tha), Turkish (\Tur), and Chinese (\Chi).
We then implement heuristics to identify entities and their translations from English Wikipedia.
Specifically, we parse the first paragraph of each Wikipedia article and pair the found translation with a topic entity of the article.
For instance, a Korean translation of \texttt{Banpo Bridge} is found from the first sentence of \url{https://en.wikipedia.org/wiki/Banpo_Bridge}.
Per-language statistics are reported in Table~\ref{tab:entity-translation-statistics}.

\begin{table*}[t]
    \centering \myfontsize
    \begin{tabular}{l @{\hspace{-0.2em}} r rrrr }
        \toprule
            Model & \#Params &  \multicolumn{2}{c}{AGN} &  \multicolumn{2}{c}{SST-2} \\
            \cmidrule(lr){3-4} \cmidrule(lr){5-6} 
            && $0$-shot & $4$-shot & $0$-shot & $4$-shot \\
        \midrule
            \multicolumn{6}{l}{\textbf{\em Baselines (Parametric)}} \\
            RoBERTa & x1.0 &                                    71.3 & - & 84.5 & - \\
            GPT-3 2.7B~\citep{zhao2021calibrate} & x7.6 &       44.7 & 43.3 & 57.2 & 59.1\\
            ~~~~+ CC~\citep{zhao2021calibrate} & x7.6 &         63.2 & 71.1 & 71.4 & 79.9 \\

            GPT-3 2.7B~\citep{holtzman2021surface} & x7.6 &     69.0 & - & 53.8 & 88.1 \\
            ~~~~+ PMI~\citep{holtzman2021surface} & x7.6 &      67.9 & - & 72.3 & 87.7 \\
            
            GPT-3 6.7B~\citep{holtzman2021surface} & x19 &      64.2 & - & 54.5 & 92.9 \\
            ~~~~+ PMI~\citep{holtzman2021surface} & x19 &       57.4 & - & 80.0 & 79.8 \\
            
            GPT-3 13B~\citep{holtzman2021surface} & x37 &       69.8 & - & 69.0 & 85.4 \\
            ~~~~+ PMI~\citep{holtzman2021surface} & x37 &       70.3 & - & 81.0 & 86.9 \\
            
            GPT-3 175B~\citep{zhao2021calibrate} & x500 &       43.9 & 61.0 & 71.6 & 93.6\\
            ~~~~+ CC~\citep{zhao2021calibrate} & x500 &         73.9 & 85.9 & 75.8 & 94.3 \\
            
            GPT-3 175B~\citep{holtzman2021surface} & x500 &     75.4 & - & 63.6 & 89.9 \\
            ~~~~+ PMI~\citep{holtzman2021surface} & x500 &      74.7 & - & 71.4 & 95.5 \\
        \midrule
            \multicolumn{6}{l}{\textbf{\em Ours (\Np)}} \\
            \ourssingle           & x1.0 & 74.2 & - & 86.8 & - \\
            \ours      & x1.0 & 74.5 & - & 87.2 & - \\
        \bottomrule
    \end{tabular}\vspace{-.3em}
    \caption{
        Comparison to GPT-3 on AG News and SST-2. 
        {\em \# Params} indicates the relative number of model parameters compared to RoBERTa large (354M).
        All GPT-3 numbers are taken from previous work.
        $k$-shot indicates that the model performs in-context learning with $k$ labeled examples with no gradient updates.
        We report on SST-2 and AGN, because they are all datasets shared between our paper and previous papers that report GPT-3 results~\citep{zhao2021calibrate,holtzman2021surface}.
        Our zero-shot models outperform 500x larger zero-shot GPT-3 and 7.6x larger 4-shot GPT-3, but lag behind 4-shot GPT-3 that is 19x or larger.
    }\label{tab:comparison-to-gpt3}
\end{table*}

\begin{figure*}[t]
\centering
\resizebox{1.5\columnwidth}{!}{\includegraphics{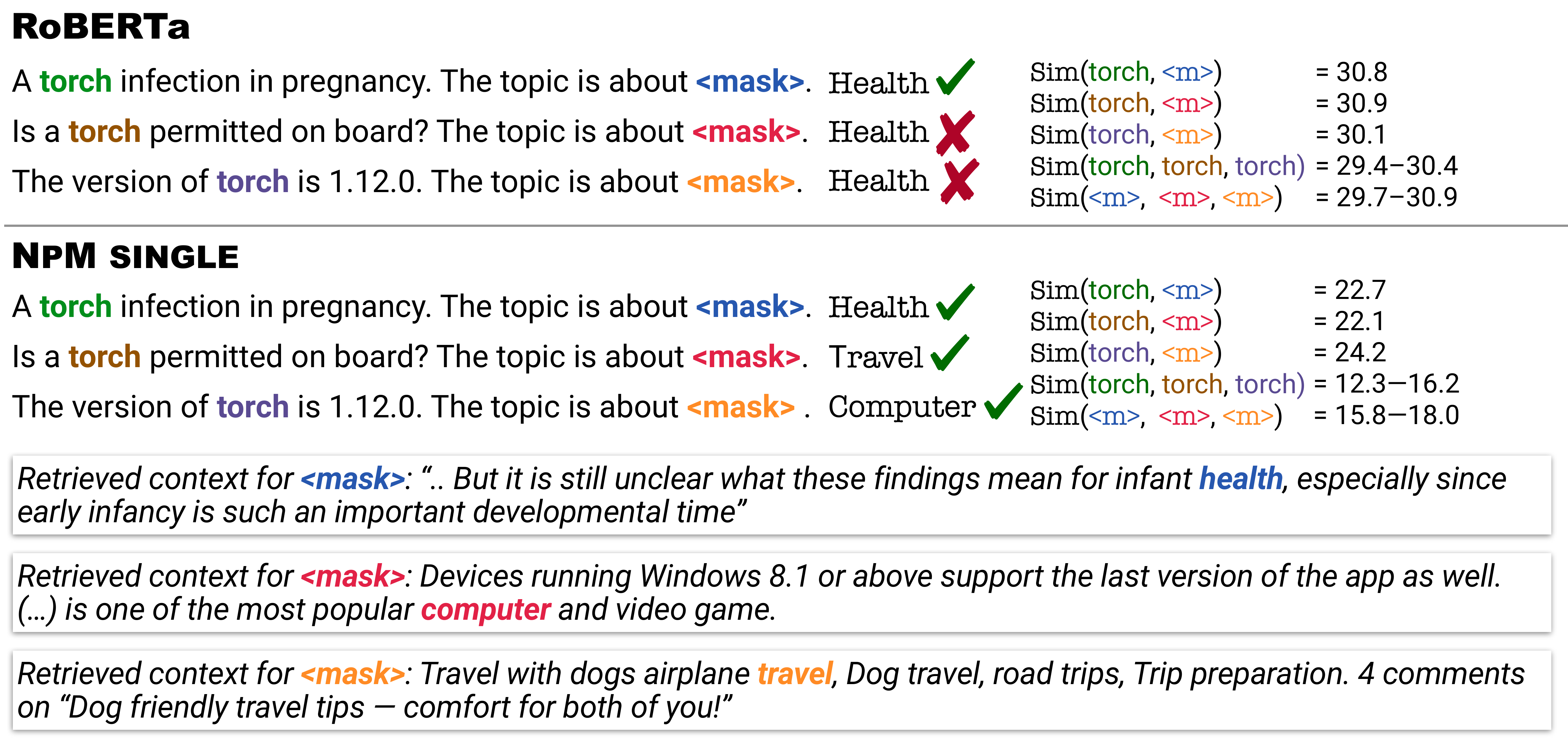}}\vspace{-.3em}
\caption{
Predictions from RoBERTa (baseline) and \ours\ on a topic classification task (classes=\{{\em health, computer, travel, politics}\}).
The bottom indicates the context \ours\ retrieves to fill in \mask.
On the right, we indicate the token-wise similarity scores.
\ours\ assigns significantly lower scores to the token pairs with distinct meanings than to the token pairs with the similar meaning, e.g., \mygreen{\em torch} ({\em a disease}) and \mybrown{\em torch} ({\em a tool}).
}\label{fig:prediction-torch}
\end{figure*}

\begin{table*}[t]
    \centering \myfontsize
    \setlength{\tabcolsep}{2.2pt}
    \begin{tabular}{l @{\hspace{-0.5em}} r @{\hspace{0.8em}} r @{\hspace{0.6em}} rrr rr rrr rrr rrr ra}
        \toprule
            Model & \#Params & \#L &
            \Ara & \Cze & \Gre & \Hin & \Heb & \Jap & \Kor & \Mal & \Mon & \Pol & \Rus & \Tal & \Tha & \Tur & \Chi & AVG \\
        \midrule
            \multicolumn{16}{l}{\textbf{\em Baselines, English-only}} \\
            T5          & 2.2x &&0.0&0.0&0.0&0.0&0.1&0.2&0.0&0.0&0.0&1.1&0.9&0.0&0.0&0.0&0.0&0.2 \\
            T5 3B       & 8.5x && 0.0 & 0.0 & 0.0&0.0&0.0&0.0&0.0&0.0&0.0&5.6&1.3&0.0&0.0&0.0& 0.0 & 0.5 \\
            OPT 6.7B    & 19x       &&0.0&0.0&0.3&0.0&0.0&0.0&3.1&0.0&0.0&0.0&2.9&0.0&0.0&0.0&0.0&0.4 \\
            OPT 13B     & 37x       &&1.5&0.0&1.2&0.7&0.0&0.0&1.4&0.0&0.0&1.1&7.4&0.0&0.0&1.3&0.1&1.0 \\
        \cmidrule(lr){1-19}
            BM25 + T5 & 2.2x &&0.0&5.5&0.3&0.2&0.5&0.0&0.2&1.9&0.0&6.8&0.8&1.2&0.0&11.3&0.0&1.9 \\
            BM25 + T5 3B & 8.5x && 0.0&12.8&0.1&0.7&0.2&0.8&0.0&0.0&1.6&28.8&1.7&0.0&0.0&20.0&0.0&4.4 \\
            BM25 + OPT 6.7B  & 19x   &&26.4&\textbf{54.1}&15.5&11.2&11.8&14.4&19.6&5.7&3.1&\textbf{47.5}&52.5&6.2&12.2&32.0&22.7&22.3 \\
            BM25 + OPT 13B  & 37x   &&17.3&51.4&24.9&15.5&27.8&12.3&22.0&11.3&7.8&45.8&48.2&8.8&18.9&\textbf{34.0}&23.3&24.6 \\
        \midrule
            \multicolumn{16}{l}{\textbf{\em Ours, English-only}} \\
            \ours  & 1.0x &&
            \textbf{51.9} & 33.0 & \best{60.9} & \textbf{63.2} & \best{63.7} & \best{59.0} & \textbf{60.5} & \best{50.9} & \textbf{46.9} & 33.3 & \best{61.2} & \best{51.2} & \best{60.8} & 32.7 & \textbf{56.9} & \textbf{52.4 }\\
        \midrule
            \multicolumn{16}{l}{\textbf{\em References, Multilingual}} \\
            mT5         & 3.4x & 101 &0.3&1.8&1.5&0.0&0.4&1.9&0.7&0.0&0.0&1.1&4.6&2.5&1.4&3.3&0.7&1.3 \\
            mT5 XL      & 11x & 101 &4.4&3.7&4.9&6.8&0.7&2.3&4.1&1.9&4.7&5.6&8.0&5.0&0.0&6.7&2.8&4.1 \\
            BLOOM 3B & 8.5x & 46 &0.0&0.0&0.0&0.0&0.0&0.0&0.0&0.0&0.0&0.0&0.0&0.0&0.0&0.0&0.3&0.0 \\
            BLOOM 7.1B & 20x & 46 &0.0&0.9&0.0&0.0&0.0&0.0&0.0&0.0&0.0&0.0&0.0&0.0&0.0&0.0&0.5&0.1 \\
        \cmidrule(lr){1-19}
            BM25 + mT5 & 3.4x & 101 &12.4&22.9&21.6&9.8&12.5&28.9&19.1&11.3&18.8&15.8&16.0&17.5&28.4&16.7&33.4&19.0 \\
            BM25 + mT5 XL & 11x & 101 &\best{64.4}&\best{64.2}&54.3&\best{65.6}&62.7&55.4&\best{69.4}&43.4&\best{62.5}&52.0&53.7&37.5&50.0&\best{48.7}&\best{65.0}&\best{56.6} \\
            BM25 + BLOOM 3B & 8.5x & 46 &24.2&25.7&1.7&13.3&15.1&18.5&17.9&5.7&6.2&21.5&11.1&10.0&27.0&18.0&44.5&17.4 \\
            BM25 + BLOOM 7.1B & 20x & 46 &19.0&49.5&11.4&20.8&8.1&30.1&25.4&5.7&6.2&\best{54.2}&29.0&6.2&37.8&33.3&53.7&26.0 \\
        \bottomrule
    \end{tabular}\vspace{-.1em}
    \caption{
        Results on the entity translation task.
        {\em \#L} indicates the number of languages multilingual models are trained on.
        \textbf{Bold} and \best{Bold} indicate the best among monolingual models and the best including multilingual models, respectively.
        {\ours\ significantly outperforms all existing monolingual models, and approaches or outperforms larger multilingual models.}
    }\label{tab:entity-translation}
\end{table*}

\begin{table*}[t]
    \centering \myfontsize
    \setlength{\tabcolsep}{2.2pt}
    \begin{tabular}{l @{\hspace{-0.5em}} r @{\hspace{0.8em}} r @{\hspace{0.6em}} rrr rr rrr rrr rrr ra}
        \toprule
            Model & \#Params & \#L &
            \Ara & \Cze & \Gre & \Hin & \Heb & \Jap & \Kor & \Mal & \Mon & \Pol & \Rus & \Tal & \Tha & \Tur & \Chi & AVG \\
        \midrule
            \multicolumn{16}{l}{\textbf{\em Baselines, English-only}} \\
            T5          & 2.2x &&0.0&13.8&0.7&0.9&0.6&1.1&0.5&3.8&1.6&15.8&1.3&7.5&0.0&16.7&0.4&4.0 \\
            T5 3B       & 8.5x &&0.2&21.1&1.0&0.7&0.7&2.3&1.2&3.8&4.7&37.3&2.9&8.8&1.4&30.7&0.4&7.3 \\
            OPT 6.7B    & 19x   &&24.4&56.9&22.9&15.5&19.7&19.1&32.5&24.5&3.1&\textbf{56.5}&60.9&22.5&23.0&46.0&30.2&30.5 \\
            OPT 13B     & 37x   &&20.7&\textbf{62.4}&22.7&15.7&30.9&17.6&36.1&18.9&15.6&\textbf{56.5}&52.2&22.5&35.1&\textbf{48.7}&40.0&33.0 \\
        \midrule
            \multicolumn{16}{l}{\textbf{\em Ours, English-only}} \\
            \ours      & 1.0x && \textbf{70.3} & 44.0 & \best{76.8} & \textbf{74.0} & \best{82.4} &\best{71.3} & \textbf{73.2} & \best{58.5} &\textbf{59.4}&45.2&\best{71.5}&\best{68.8}&\textbf{66.2}&45.3&\textbf{74.5} & \textbf{65.4} \\
        \midrule
            \multicolumn{16}{l}{\textbf{\em References, Multilingual}} \\
            mT5         & 3.4x & 101    &19.4&25.7&30.8&19.0&20.6&33.8&28.2&28.3&40.6&18.6&23.1&30.0&29.7&26.7&37.4&27.5 \\
            mT5 XL      & 11x & 101 &\best{83.2}&\best{76.1}&69.6&\best{81.5}&77.4&68.2&\best{85.2}&49.1&\best{67.2}&\best{65.5}&62.7&51.2&\best{68.9}&\best{64.0}&\best{79.0}&\best{69.9} \\
            BLOOM 3B & 8.5x & 46        &51.2&27.5&3.1&30.2&34.1&34.0&30.9&11.3&7.8&28.2&23.0&17.5&37.8&22.0&70.1&28.6 \\
            BLOOM 7.1B & 20x & 46       &29.6&43.1&12.0&27.6&12.2&32.5&30.9&9.4&15.6&59.3&38.1&13.8&43.2&32.0&65.5&31.0 \\
        \bottomrule
    \end{tabular}\vspace{-.3em}
    \caption{
        Results on the entity translation task given an oracle passage.
        {\em \#L} indicates the number of languages multilingual models are trained on.
        \textbf{Bold} and \best{Bold} indicate the best excluding multilingual models and the best including multilingual models, respectively.
    }\label{tab:entity-translation-oracle}
\end{table*}

\section{Additional Results}\label{app:additional-results}\paragraph{Full results on knowledge tasks.}
Table~\ref{tab:open-set} reports full results on five knowledge tasks. See Figure~\ref{fig:open-set-results} for an illustration, and Section~\ref{subsec:results-open} for discussion.

\vspace{-.2em}
\paragraph{Comparison to few-shot GPT-3.}
Table~\ref{tab:comparison-to-gpt3} compares zero-shot \ourssingle\ and \ours\ with zero- and four-shot GPT-3.
Our zero-shot models outperform 500x larger zero-shot GPT-3 and 7.6x larger 4-shot GPT-3, but lag behind 4-shot GPT-3 that is 19x or larger.
We think future work can explore extending our models to a few-shot setup.

\vspace{-.2em}
\paragraph{Additional qualitative results.}
Figure~\ref{fig:prediction-torch} depicts predictions from RoBERTa and \ours\ in topic classification, choosing a label between four candidates: {\em health}, {\em computer}, {\em travel} and {\em politics}.
All three examples contain the word {\em torch}, but with different meanings, e.g., an infectious diseases, a tool, and a computer library.
RoBERTa predicts {\em health} for all of them, while \ours\ predicts {\em health}, {\em travel} and {\em computer}, which are all correct predictions.

As in Figure~\ref{fig:prediction}, we find that representations from \ours\ enable better word sense disambiguation: the pairwise similarities between between different meanings of {\em torch} are significantly lower than the pairwise similarities between other tokens that share the meaning.

\vspace{-.2em}
\paragraph{Entity translation given an oracle passage.}
We evaluate models on the entity translation task where an oracle passage---a passage that is guaranteed to contain the translation information---is provided to the model.
Baselines prepend oracle passages to the input, as it does with the retrieve-and-generate approach.
\ours\ uses oracle passages to restrict the search space.

Table~\ref{tab:entity-translation-oracle} reports results. While performance overall increases compared to when the oracle passage is not provided, the overall comparison between models does not change from Table~\ref{tab:entity-translation}: (1) all monolingual models significantly suffer, except for a couple of languages that are derived from Latin; (2) \ours\ significantly outperforms all monolingual models; (3) \ours\ even outperforms 3.4x larger mT5 and 20x larger BLOOM, and approaches 11x larger mT5.

\end{document}